\documentclass[runningheads]{llncs}

\usepackage{graphicx}
\usepackage{amsmath,amssymb,amsthm}
\usepackage{mathrsfs}
\usepackage{url}
\usepackage[]{algorithm2e}
\usepackage[]{subfigure}

\usepackage{xcolor}

\begin{document}
\title{Abstraction of Markov Population Dynamics via Generative Adversarial Nets}

\author{Francesca Cairoli\inst{1} \and
Ginevra Carbone\inst{1}\and
Luca Bortolussi\inst{1,2}}
\authorrunning{F. Cairoli et al.}

\institute{Department of Mathematics and Geosciences, University of Trieste, Italy \and
Modeling and Simulation Group, Saarland University, Germany\\}
\maketitle              
\begin{abstract}
Markov Population Models are a widespread formalism used to model the dynamics of complex systems, with applications in Systems Biology and many other fields. The associated Markov stochastic process in continuous time is often analyzed by simulation, which can be costly for large or stiff systems, particularly when a massive number of simulations has to be performed (e.g. in a multi-scale model). A strategy to reduce computational load is to abstract the population model, replacing it with a simpler stochastic model, faster to simulate. Here we pursue this idea, building on previous works and constructing a generator capable of producing stochastic trajectories in continuous space and discrete time. This generator is learned automatically from simulations of the original model in a Generative Adversarial setting. Compared to previous works, which rely on deep neural networks and Dirichlet processes, we explore the use of state of the art generative models, which are flexible enough to learn a full trajectory rather than a single transition kernel. 
\end{abstract}

\vspace{-0.35cm}

\section{Introduction}\label{sec:introduction}

A wide range of complex systems can be modeled as a network of chemical reactions. Stochastic simulation is typically the only feasible analysis approach that scales in a computationally tractable manner with the increase in system size, as it avoids the explicit construction of the state space. The well known Gillespie Stochastic Simulation Algorithm~\cite{gillespie1977exact} is widely used for simulating models, as it  samples from the exact distribution over trajectories. This algorithm is  effective to simulate systems of moderate complexity, but it does not scale well to systems with many species and reactions, large populations, or internal stiffness. In these scenarios, a more effective choice is to rely on approximate simulation algorithms such as tau-leaping~\cite{cao2006efficient} and hybrid simulation~\cite{pahle2009biochemical}. 
Nonetheless, when the  number of simulations required is extremely large and possibly costly, e.g. when one needs to simulate a large population of heterogeneous cells in a multi-scale model of a tissue or to simulate many heterogeneous individuals in an population ecology scenario,  all these methods become extremely computationally demanding, even for HPC facilities. 

A viable approach to address such problem is model abstraction, which aims at reducing the underlying complexity of the model, and thus reduce its simulation cost. However, building effective model abstractions is difficult, requiring a lot of ingenuity and man power. Here we advocate the strategy of learning an abstraction from simulation data. Our strategy is to frame model abstraction as a supervised learning problem, and learn an abstract probabilistic model using state of the art deep learning. 
The probabilistic model should then be able to generate approximate trajectories efficiently and in constant time, i.e., independent on the complexity of the original system, thus sensibly reducing the simulation cost.
\paragraph{Related work.}
The idea of using machine learning as a model abstraction tool to approximate and simplify the dynamics of a Markov Population Process has received some attention in recent years. 
In~\cite{bortolussi2018deep} the authors use a Mixture Density Network (MDN)~\cite{bishop2006pattern} to approximate the transition kernel of the stochastic process. In~\cite{petrov2020automated} the authors extend the previous approach by introducing an automated search of the MDN architecture that better fit the data. In~\cite{bortolussi2019bayesian} the authors present a Bayesian model abstraction technique, based on Dirichlet Processes, that allows  the quantification of the reconstruction uncertainty. In all cases, what is learned is an approximate transition kernel,  i.e., the probabilistic distribution of a single simulation step. 

In this paper we address a more general and more complex problem. Instead of learning an approximate transition kernel, we learn the distribution of an entire trajectory of fixed length. This latter problem is not solvable with any of the previously adopted approaches, and its major goal is to keep abstraction error under control. In fact, training the abstract model on a full trajectory, rather than on pairs of subsequent states, allows the abstract model to retain and capture more information about the dynamics of the Markov process.

\paragraph{Contributions.}
Our approach leverages Generative Adversarial Nets (GAN), which are one of the most strong and flexible techniques to learn probabilistic models. In fact, the GAN-based model abstraction technique is capable of learning a conditional distribution over the trajectory space, keeping into account the correlation, both spatial and temporal, among all the different species and conditioning both on initial states and model parameters. All the previous approaches focus on learning the distribution of the state of the system after a time $\Delta t$, the so called \textit{transition kernel}. However, such approaches perform poorly when the time interval is small and the dynamics is transient, showing a clear propagation of the error as the approximate kernel is applied iteratively to form a trajectory. 
Furthermore, producing a full trajectory reduces even more the computational cost of simulating a large pool of trajectories for different initial settings.

\paragraph{Paper structure.}
The paper is organized as follows: in Section~\ref{sec:background} the relevant background notions are introduced, in Section~\ref{sec:abstraction} we describe in detail the abstraction procedure, Section~\ref{sec:experiments} presents the case studies and the experimental evaluation. 
Conclusions are drawn in Section~\ref{sec:conclusions}.

\vspace{-0.25cm}

\section{Background}\label{sec:background}

\subsection{Chemical Reaction Networks}
 
Consider a system with $n$ species 
evolving according to a stochastic model defined as a Chemical Reaction Network. Under the well-stirred assumption, the time evolution can be modelled as a Continuous Time Markov
Chain (CTMC) on a discrete state space. 
The vector $\eta_t = (\eta_{t,1},\dots,\eta_{t,n})\in S\subseteq\mathbb{N}^n$ denotes the state vector at time $t$, where $\eta_{t,i}$ is the number of individuals in species $i$ at time $t$. The dynamics is encoded by a set of $m$ reactions with parametric propensity functions that depends on the state of the system. 
Due to the memoryless property of CTMC, 
the probability of finding the
system in state $s$ at time $t$ given that it was in state $s_0$ at time $t_0$ can be expressed as a system of ODEs known as Chemical Master Equation (CME). 
Since in general the CME is a system with countably many differential equations, its analytic or numeric solution is almost always unfeasible. An alternative
computational approach is to generate trajectories using stochastic algorithms
for simulation, like the well-known Gillespie’s SSA~\cite{gillespie1977exact} which produces statistically correct trajectories, i.e., sampled according to the stochastic process described by the CME.

\vspace{-0.2cm}

\subsection{Generative Adversarial Nets}\label{sec:gan_intro}

Every dataset can be considered as a set of observations drawn from an unknown distribution $\mathbb{P}_r$. Generative models aim at learning a model that mimics this unknown distribution as closely as possible, i.e., learn a distribution $\mathbb{P}_{w_g}$ as similar as possible to $\mathbb{P}_r$, in order to then get samples from it that are new but look as if they could have belonged to the original dataset.
Generative Adversarial Nets (GANs)~\cite{goodfellow2014generative} are deep learning-based generative models, that, given a dataset, are capable of generating new random but plausible examples. 

\paragraph{Wasserstein GAN.} In this work we consider the Wasserstein version of GAN (WGAN)~\cite{arjovsky2017wasserstein,gulrajani2017improved} as it is known to be more stable and less sensitive to the choice of model architecture and hyperparameters compared to a traditional GAN.
WGANs use the Wasserstein distance (also known as Earth-Mover's distance), rather than the Jensen Shannon divergence, to measure the difference between the model distribution $\mathbb{P}_{w_g}$ and the target distribution $\mathbb{P}_r$. Because of  Kantorovich-Rubinstein duality \cite{villani2008optimal} such distance can be computed as the supremum over all the 1-Lipschitz functions $f : S \rightarrow \mathbb{R}$:
\begin{equation}\label{eq:wassdist}
\small 
W(\mathbb{P}_r,\mathbb{P}_{w_g}) =  \sup_{||f||_L\le 1} \left(
\mathbb{E}_{x\sim\mathbb{P}_{r}}[f(x)]-\mathbb{E}_{x\sim\mathbb{P}_{w_g}}[f(x)]
\right).
\end{equation}
We approximate these functions $f$ with a neural net $C_{w_c}$ parametrized by weights $w_c$. 
To enforce the Lipschitz constraint we follow \cite{gulrajani2017improved} and introduce a penalty over the norm of the gradients. It is known that a differentiable function is 1-Lipchitz if and only if it has gradients with norm at most 1 everywhere. 
The objective function, to be maximized w.r.t. $w_c$, becomes:
\begin{equation}\label{eq:wassdist_gp}
\small 
\mathcal{L}({w_c}, w_g) := \mathbb{E}_{x\sim\mathbb{P}_{r}}[C_{w_c}(x)]-\mathbb{E}_{x\sim\mathbb{P}_{w_g}}[C_{w_c}(x)]-\lambda \mathbb{E}_{\hat{x}\sim\mathbb{P}_{\hat{x}}}
( \lVert \nabla_{\hat{x}}C_{w_c} (\hat{x})  \lVert_2-1 )^2]  ,
\end{equation}
where $\lambda$ is the penalty coefficient and $\mathbb{P}_{\hat{x}}$ is defined by sampling uniformly along straight lines between pairs of points sampled from $\mathbb{P}_r$ and $\mathbb{P}_{w_g}$. This is actually a softer constraint that however performs well in practice~\cite{gulrajani2017improved}.
The $C_{w_c}$ network is referred to as \textit{critic} and it outputs different scores for real and fake samples, its objective function (Eq.~\eqref{eq:wassdist_gp}) provide an estimate of the Wasserstein distance among the two distributions. On the other hand, the distribution $\mathbb{P}_{w_g}$ is parametrized by $w_g$; we seek the parameters that make it as close as possible to $\mathbb{P}_r$. To achieve this, we consider a random variable $Z$ with a fixed simple distribution $\mathbb{P}_Z$ and pass it through a parametric function, the \textit{generator}, $G_{w_g} : Z \rightarrow S$ that
generates samples following the distribution $\mathbb{P}_{w_g}$.
Therefore, the WGAN architecture consists of two deep neural nets, a generator that proposes a distribution and a critic that estimate the distance between the proposed and the real (unknown) distribution. Using WGAN brings several important advantages compared to traditional GAN: it avoids the mode collapse problem, which makes WGAN more suitable for capturing stochastic dynamics, it drastically reduces the problem of vanishing gradients and it also have an objective function that correlates with the quality of generated samples, making the results easier to interpret. 

\paragraph{Conditional GAN.} Conditional Generative Adversarial Nets (cGAN)~\cite{mirza2014conditional} are a type of GANs that involves the conditional generation of examples, i.e., the generator produces examples of a required type, e.g. examples that belong to a certain class, and thus they introduce control over the desired generated output. In our application, we want the generation of stochastic trajectories to be conditioned on some model parameters and on the initial state of the system.

Furthermore, dealing with inputs that are trajectories, i.e. sequences of fixed length, requires the use of convolutional neural networks (CNNs)~\cite{goodfellow2016deep} for both the generator and the critic. 
The architecture used in this work is thus a conditional Wasserstein Convolutional GAN with gradient penalty, it is going to be referred to as cWCGAN-GP.

\vspace{-0.25cm}

\section{GAN-based Abstraction}\label{sec:abstraction}

\subsection{Model Abstraction}

The underlying idea is the following: given a stochastic process $\{\eta_{t}\}_{t\ge 0}$ with transition probabilities $\mathbb{P}_{s_0}(\eta_{t}=s) = \mathbb{P}(\eta_{t}=s\mid \eta_{t_0}=s_0)$, we aim at finding another stochastic process whose trajectories are faster to simulate but similar to the original ones. 
Time has to be discretized, meaning we fix an initial time $t_0$ and a time step $\Delta t$ that suits our problem. We define $\tilde{\eta}_i := \eta_{t_0+i\cdot\Delta t}$, $\forall i\in\mathbb{N}$. In addition, 
given a fixed time horizon $H$, we define time-bounded trajectories as $\tilde{\eta}_{[1,H]} = s_1 s_2\cdots s_H\in S^H\subseteq\mathbb{N}^{H\times n}$. Given a state $s_0$ and a set of parameters $\theta$, we can represent a trajectory of length $H$ as a realization of a random variable over the state space $S^H$. 
The probability distribution for such random variable is given by the product of the transition probabilities at each time step: 
${ \mathbb{P}_{s_0,\theta}(\tilde{\eta}_{[1,H]}=s_1s_2\cdots s_H)=\prod_{i=1}^H\mathbb{P}_{s_{i-1}, \theta}(\tilde{\eta}_{i}=s_{i})}$.
The CTMC, $\{\eta_{t}\}_{t\ge 0}$, is now expressed as a time-homogeneous Discrete Time Markov Chain $\{\tilde{\eta}_i\}_i$. 
An additional approximation has to be made: the abstract model takes values in $S'\subseteq\mathbb{R}_{\geq 0}^n$, a continuous space in which the state space $S\subseteq\mathbb{N}^n$ is embedded.
In constructing the approximate probability distribution for trajectories we can decide to restrict our attention to arbitrary aspects of the process, rather than trying to preserve the full behavior. A \emph{projection} $\pi$ from $S^H$ to an arbitrary space $U^H$ can be used to reach this purpose,  for instance, to monitor the number of molecules belonging to a certain subset of chemical species, i.e., $U\subseteq S$. 
Note that $\pi(\tilde{\eta}_{[0,H]})$ is a random variable over $U^H$. 
Such flexibility could be extremely helpful in capturing the dynamics of systems in which some species are not observable.

\paragraph{Abstraction accuracy.} Another important ingredient is a meaningful quantification of the error introduced by the abstraction procedure, i.e., the reconstruction accuracy. Such quantification must be based on a distance, $d$, among distributions. 
We choose the Wasserstein distance, together with the absolute and relative difference among means and variances of the histograms.
Given a distribution over initial states $s_0$ and a distribution over parameters $\theta$, we would like to measure the expected error at every time instant $t_i = t_0+i\cdot\Delta t$ with $i\in\{1,\dots,H\}$. Formally, we want to measure $\mathbb{E}_{s_0,\theta}\left[d\big(\pi(\eta_{[1,H]})\big |_i, \pi'(\eta'_{[1,H]})\big |_i\big)\right]$ where $\pi(\eta_{[1,H]})\big |_i$ denotes the $i$-th time components of the projected trajectory $\pi(\eta_{[1,H]})\in U^H$. 
To estimate such quantity we use a well-known unbiased estimator, which is the average over the distances computed over a large sample set of  initial settings. Computing the distance among SSA and abstract distributions at each time step quantifies how small the expected error is and, more importantly, 
how it evolves in time. As a matter of fact, it shows whether the error tends to propagate or not and how much each species contributes to the abstraction error. 
In practice, we compute $H \cdot n$ distances among distributions over $\mathbb{N}$ as we want to know how each species contributes in the reconstruction error. 

\vspace{-0.2cm}

\subsection{Dataset Generation}

\paragraph{Training set.} 
Choose a set of $N_{train}$ initial settings and for each setting simulate $k_{train}$ SSA trajectory of length $H$. The training set is composed of $N_{train}\cdot k_{train}$ pairs initial setting-trajectory, i.e. pairs $(\theta^i,s_0^i,\eta^{ij}_{[1,H]})$ for $i=1, \ldots,N_{train}$ and $j= 1,\ldots, k_{train}$.

\paragraph{Test set.} 

Choose a set of $N_{test}$ initial settings and for each setting simulate a large number, $k_{test}\gg k_{train}$, of SSA trajectory of length $H$. The test set is composed of $N_{test}\cdot k_{test}$ pairs initial setting-trajectory, i.e. pairs $(\theta^i,s_0^i,\eta^{ij}_{[1,H]})$ for $i=1, \ldots,N_{test}$ and $j= 1,\ldots, k_{test}$. 

\paragraph{Partial observability.} In case of partial observability, $U\subseteq S$,  we fix an initial condition for species in $U$, and simulate a pool of trajectories each time sampling the initial value of species in $S\smallsetminus U$. As a result, we are learning and abstract distribution that marginalizes over unobserved variables.

\vspace{-0.2cm}

\subsection{cWCGAN-GP architecture} 

The critic $C_{w_c}$ takes as input a batch of initial states, $s_0^1,\dots , s_0^b$, a batch of parameters, $\theta_1, \dots , \theta_b$, and a batch of subsequent trajectories, $\eta_{[1,H]}^1,\dots , \eta_{[1,H]}^b$. For each $i\in\{ 1,\dots , b\}$ the inputs, $\eta_{[1,H]}^i$, $s_0^i$ and $\theta_i$, are concatenated to form an input with dimension $b\times (H+1)\times (n+m)$. 
Formally, $C_{w_c}: S^{H+1}\times\Theta\rightarrow \mathbb{R}$.  To enforce the Lipschitz property over $C_ {w_c}$ we add a gradient penalty term over $\mathbb{P}_{\hat{x}}$. Samples of $\mathbb{P}_{\hat{x}}$ are generated by sampling uniformly along straight lines connecting points coming from a batch of real trajectories and points coming from a batch of generated trajectories.

On the other hand, the generator $G_{w_g}$ takes as input a batch of initial states, $s_0^1,\dots , s_0^b$, a batch of parameters, $\theta_1, \dots , \theta_b$, and a batch of random noise, $z^1,\dots , z^b$, with dimension $k$, a user-defined hyper-parameter. For each $i\in\{ 1,\dots , b\}$ the two inputs are, once again, concatenated to form an input with dimension $b\times (n+m+k)$. The generator outputs a batch of generated trajectories $\hat{\eta}_{[1,H]}^1,\dots\hat{\eta}_{[1,H]}^b$. Formally, $G_{w_g}:S\times\Theta\times Z\rightarrow S^H$, such that $G_{w_g}(s_0,\theta,z) = \hat{\eta}_{[1,H]} = s_1\cdots s_H$. See the pseudocode for the algorithm in Appendix~\ref{sec:algorithm} of [XXX].

\vspace{-0.2cm}

\subsection{Model Training}
The cWCGAN-GP-based model abstraction framework consists in training two different CNNs.
The loss function, introduced in Eq.~\eqref{eq:wassdist_gp}, is a parametric function depending both on the generator weights $w_g$ and the critic weights $w_c$. 
When training the critic, we keep the generator weights constant $\overline{w}_g$, and we maximize $\mathcal{L}(w_c, \overline{w}_g)$ w.r.t. $w_c$.  
Formally, we solve the problem 
\begin{equation*}
   \small
   w_c^* = \underset{w_c}{\mbox{argmax}}\Big\{\mathcal{L}(w_c, \overline{w}_g)\Big\}. 
\end{equation*}
On the other hand, in training the generator, we keep the critic weights constant $\overline{w}_c$, 
and we minimize $\mathcal{L}(\overline{w}_c, w_g)$ w.r.t. $w_g$. 
Formally, we solve the problem 
\begin{equation*}
    \small
    w_g^* = \underset{w_g}{\mbox{argmin}} \Big\{\mathcal{L}(\overline{w}_c,w_g)\Big\} =  \underset{w_g}{\mbox{argmin}}  \Big\{ -\mathbb{E}_{z,(s_0,\theta)}\Big[C_{\overline{w}_c}\big(G_{w_g}(z,s_0,\theta),s_0,\theta)\Big]\Big\}.
\end{equation*}
As mentioned in Section \ref{sec:background}, the loss function derives from the Wasserstein distance between the real and generated distributions, see \cite{arjovsky2017wasserstein,gulrajani2017improved} for the mathematical details.

Intuitively, the generator generates a batch of samples, and these, along with real examples from the dataset, are provided to the critic, which is then updated to get better at estimating the distance between the real and the abstract distribution. The generator is then updated based on scores obtained by the generated samples from the critic. An important collateral advantage is that WGANs have a loss function that correlates with the quality of generated examples.

Training the cWCGAN-GP has a cost. Nonetheless, once it has been trained, its evaluation is extremely fast. Details about training and evaluation costs are discussed in Section~\ref{sec:experiments}.

\paragraph{Abstract Model Simulation.}
Once the training is over, we can discard the critic and focus only on the trained generator $G$. In order to generate an abstract trajectory starting from a state $s_0^*$ with parameters $\theta^*$, we just have to sample a value $z$ from the random noise variable $Z$ and evaluate the generator on the pair $(s_0^*, \theta^*, z)$. The output is a stochastic trajectory of length $H$: $G(s_0^*, \theta^*, z) = \hat{\eta}_{[1,H]}$. The stochasticity is provided by the random noise variable, de facto the generator acts as a distribution transformer that maps a simple random variable  into a complex distribution. In order to generate a pool of $p$ trajectories, we simply sample $p$ different values from the random noise variable: $z_1, \dots,z_p$.
Therefore, the generation of a trajectory has a fixed computational cost. 

\vspace{-0.25cm}

\section{Experimental Results}\label{sec:experiments}

In this section we validate our GAN-based model abstraction procedure on the following case studies. More details  are provided in Appendix~\ref{sec:casestudies}.
\begin{itemize}
    \item \textbf{SIR Model (Absorbing state).} 
The SIR epidemiological model describes the spread, in a population, of an infectious disease that grants immunity to those who recover from it. The population is divided in three mutually exclusive groups: susceptible (S), infected (I) and recovered (R). The possible reactions, given by the interaction of individuals are infection and recovery. An important feature is the presence of an absorbing states.
    \item \textbf{Ergodic SIRS Model.} 
    A SIR model in which the population is not perfectly isolated, meaning there is always a chance of getting infected from some external individuals, and in which immunity is only temporary. As a consequence, this model has no absorbing state.
    \item \textbf{Genetic Toggle Switch Model (Bistability).} 
The toggle switch is a well-known bistable biological circuit consisting of two genes, $G_1$ and $G_2$, that mutually repress each other in the production of proteins $P_1$ and $P_2$ respectively. The system displays two stable equilibria.

    \item \textbf{Oscillator Model.} 
The 
circuit consists of three species A, B and C and three cyclic reactions: A converts B to itself, B converts C to itself, and C converts A to itself.
The concentrations of the three species oscillates in time.

\item \textbf{MAPK Model.} The mitogen-activated protein kinase cascade models the amplification of an output signal ($\it MAKP\_PP$) thorough a multi-level cascade with negative feedback which is ultra-sensitive to an input stimulus ($V_1$).
The output signal shows either stable or oscillating behaviour, depending on the input signal.

\end{itemize}

In order to evaluate the performance of our abstraction procedure we consider two important measures: the accuracy of the abstract model, evaluated for each species at each time step of the time grid, and the computational gain compared to SSA simulation time. 

\paragraph{Experimental Settings.} 
The workflow can be divided in steps: (1) define a CRN model, (2) generate the synthetic datasets via SSA simulation, (3) learn the abstract model by training the cWCGAN-GP and, finally, (4) evaluate such abstraction. All the steps have been implemented in Python. 
In particular, CRN models are defined in the \texttt{.psc} format, CRN trajectories are simulated using Stochpy~\cite{maarleveld2013stochpy} (stochastic modeling in Python) and PyTorch~\cite{paszke2017automatic} is used to craft the desired architecture for the cWCGAN-GP and to evaluate the latter on the test data. All the experiments were performed on a  
Intel Xeon Gold 6140 
with 24 cores and a 128GB RAM.
The source code for all the experiments can be found at the following link: \url{https://github.com/francescacairoli/WGAN_ModelAbstraction}.

\paragraph{Datasets.}
For each case study with fixed parameters, the training set consists of $20$K different SSA trajectories. In particular, $N_{train} =2$K and $k_{train}= 10$. 
The test set, instead, consists of $25$ new initial settings 
and from each of these we simulate $2$K trajectories, so to obtain an empirical approximation of the distribution targeted by model abstraction.
When a parameter is allowed to vary, the training set consists of $50$K SSA trajectories ($N_{train} =1$K and $k_{train}= 50$). 
We manually choose $H$ and $\Delta t$ so that the system is close to steady state at time $H\cdot \Delta t$, without spending there too many steps. The time interval should be small enough to capture the full transient behavior of the system. For systems with no steady state, such as the oscillating models, we choose $H$ and $\Delta t$ so to observe a full period of oscillation. The chosen values are the following: SIR: $\Delta t = 0.5$, $H = 16$; e-SIRS: $\Delta t = 0.1$, $H = 32$; Toggle Switch: $\Delta t = 0.1$, $H = 32$; Oscillator: $\Delta t = 1$, $H = 32$; MAPK: $\Delta t = 60$, $H = 32$.

\paragraph{Data Preparation.} 
Data have been scaled to the interval $[-1,1]$ to enhance the performance of the two CNNs and to avoid sensitivity to different scales in species counts. During the evaluation phase, the trajectories have been scaled back. Hence, results and errors are shown in the original scale.

\vspace{-0.2cm}

\subsection{cWCGAN-GP architecture}

The same architecture and the same set of hyper-parameters works well for all the analyzed case studies, showing great stability and usability of the proposed solution.
The Wasserstein formulation of GANs, with gradient penalty 
, strongly contributes to such stability. Traditional GANs have been tested as well, but they do not have such strength.
The details of the archictecture follows the best practice suggestions provided in~\cite{gulrajani2017improved}. 
The critic network has two hidden one-dimensional convolutional layers, with $n+m$ channels, each containing $64$ filters of size $4$ and stride $2$. We use a leaky-ReLU activation function with slope $0.2$, we do layer normalization and at each layer we introduce a dropout with probability $0.2$. An additional dense layer, with linear activation function, is used to connect the single output node, that contains the critic value. In order to enforce the Lipschitz constraint on the critic’s model we add a gradient penalty term, as described in Section~\ref{sec:gan_intro}.
On the other hand, the generator network takes as input the noise and the initial settings and it embeds the inputs in a larger space with $N_{ch}$ channels ($512$ in our experiments) through a dense layer. Four one-dimensional convolutional transpose layers are then inserted, containing respectively $128$, $256$, $256$ and $128$ filters of size $4$ with stride $2$. Here we do batch normalization and use a leaky-ReLU activation function with slope $0.2$.
Finally, a traditional convolutional layer is introduced to reduce the number of output channels to $n$. 
The Adam algorithm~\cite{bengio2015rmsprop} is used to optimize loss function of both the critic and the generator. The learning rate is set to $0.0001$ and $\beta= \{0.5,0.9\}$. The above settings are shared by all the case studies, the only exception is the more complex MAPK model for which a deeper cWCGAN-GP architecture is selected: a critic with five layers, each containing $256$ filters of size $4$ and stride $2$, and a generator with five layers, containing respectively $128$, $256$, $512$, $256$ and $128$ filters of size $4$ with stride $2$.

Training times depend on the dimension of the dataset, on the size of mini-batches, on the number of species, and on the architecture of the cWCGAN-GP. The latter has been kept constant for all the case studies. Batches of $256$ samples have been used and the number of epochs varies from $200$ to $500$ depending on the complexity of the model. Moreover, each training iteration of the generator correspond to $5$ iterations of the critic, to balance the power of the two player. The average time required for each training epoch is around one minute. 
Therefore, training the cWCGAN-GP model for $500$ epochs takes around $8$ hours leveraging the GPU.

\vspace{-0.2cm}

\subsection{Results}

\paragraph{Computational gain.}
The time needed to generate abstract trajectories does not depend on the complexity of the original system. Moreover, as the cWCGAN-GP architecture is shared by all the case studies, the computational time required to generate abstract trajectories is the same for all the case studies. In particular, considering a noise variable of size $480$, it takes around $1.75$ milliseconds (ms) to simulate a single trajectory. However, when generating batches of at least 200 trajectories the overhead reduces and the time to generate a single trajectory stabilizes around 0.8 ms. The same does not hold for the SSA trajectories, whose computational costs depends on the complexity of the model and on the chosen reaction rates. In the case studies considered the time required to simulate a single trajectory varies from $0.04$ to $0.22$ seconds, but it easily increases for more complex models or for smaller reaction rates, whereas the cost of abstract simulation stays constant.
Details about the computational gain for each model are presented in Table~\ref{table:comp_times}. Computations are performed exclusively on a single CPU processor, to perform a fair comparison. However, the evaluation of cWCGAN-GP can be further sped up using GPUs, especially for large batches of trajectories, but this would have introduced a bias in their favour. It is important to stress how GPU parallelization is extremely straightforward in PyTorch and how the time to generate a single trajectory decrease to 1.9 $\times 10^{-5}$ seconds when generating a batch of at least $2K$ trajectories (see last line of Table~\ref{table:comp_times}).

The training phase introduces a fixed overhead that affects the overall computational gain. For instance, the training phase of the MAPK model takes around 8 hours, which is equivalent to the time needed to generate 140K SSA trajectories. It follows that, together with the trajectories needed to generate the training set, the cost of the training procedure is paid off when we simulate at least 200K trajectories. In a typical biological multi-scale scenario in which we seek to simulate the evolution in time of a tissue containing millions of cells, having additional internal pathways, the number of trajectories needed for the training phase becomes negligible and the training time is soon paid off.

\begin{table}[!t]
\centering
\begin{tabular}{ |l|l|l|l|l|l| }
\hline
Model & SIR  & e-SIRS & Switch & Osc. & MAPK \\ 
\hline
SSA (direct) &  0.043 s &  0.047 s  & 0.041 s & 0.042 s& 0.224 s\\ 

- CPU (avg over 200): & 0.024 s & 0.024 s &  0.020 s  & 0.021 s& 0.211 s\\
\hline
SSA ($\tau$-leaping) & 0.054 s & 0.052 s  & 0.044 s & 0.042 s& 0.26 s \\ 

- CPU (avg over 200): & 0.018 s & 0.028 s & 0.024 s  & 0.021 s & 0.24 s\\
\hline
cWCGAN-GP & 0.00175 s & 0.00175 s & 0.00175 s  & 0.00175 s& 0.00175 s\\
- CPU (avg over 200): & 0.0008 s & 0.0008 s & 0.0008 s & 0.0008 s & 0.0008 s\\

- GPU (avg over 2000): $10^{-5}\times $ & 1.9  s& 1.9  s& 1.9  s& 1.9 s& 1.9  s\\
\hline

\end{tabular}

\caption{Comparison of the average computational time required to simulate a single trajectory either via SSA (both direct or approximate methods) or via the cWCGAN-GP abstraction. 200 trajectories are needed to reduce the CPU (single processor) overhead, whereas, 2000 trajectories are required for the on GPU overhead. \vspace{-0.8cm}}\label{table:comp_times}
\end{table}

\paragraph{Measures of performance.}
Results are presented as follows. For each model, we present a small batch of trajectories, both real and abstract. From the plots of such trajectories we can appreciate if the abstract trajectories are similar to real ones and if they capture the most important macroscopic behaviors. We also show the histograms of empirical distributions at time $t_H$ for each species to quantify the behavior over all the $2$K trajectories present in the test set (see Fig.~\ref{fig:sir_trajectories}-~\ref{fig:mapk_trajectories}). Additional plots are shown in Appendix~\ref{sec:plots} (Fig.~\ref{fig:sir_trajectories_extra}-~\ref{fig:clock_trajectories_extra}). 


\begin{figure}[ht]
    \centering
    \subfigure[eSIRS]{   \includegraphics[scale=0.235]{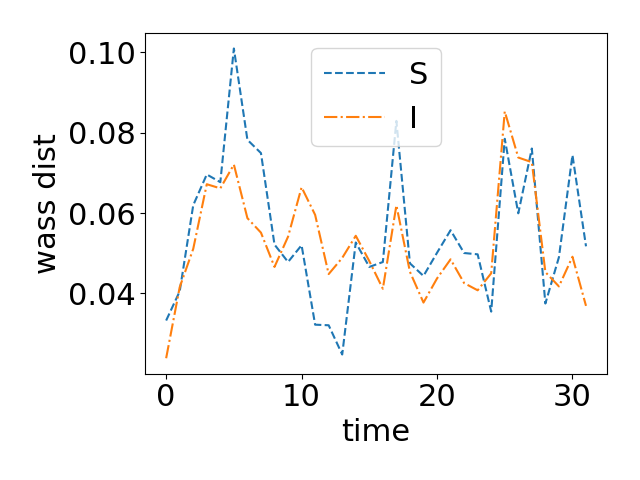}}
        \hfill
    \subfigure[eSIRS-1P]{  
    \includegraphics[scale=0.235]{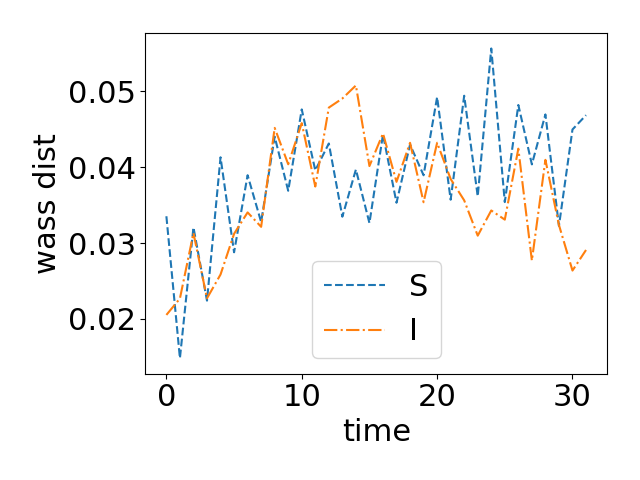}}
        \hfill
    \subfigure[SIR]{  
    \includegraphics[scale=0.235]{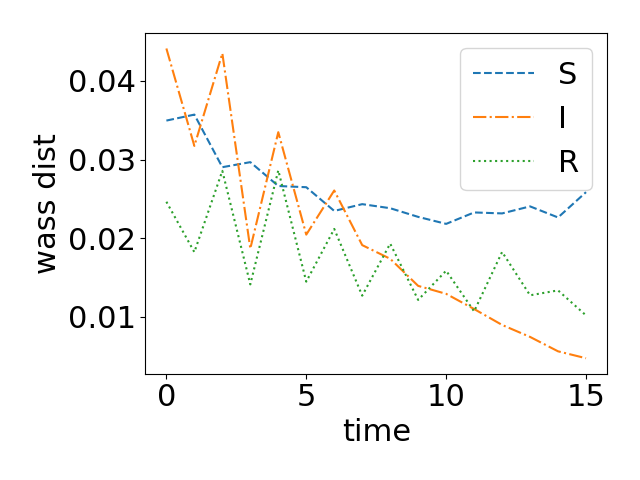}}
    
    \vspace{-0.25cm}
    
    \subfigure[Oscillator]{  
    \includegraphics[scale=0.235]{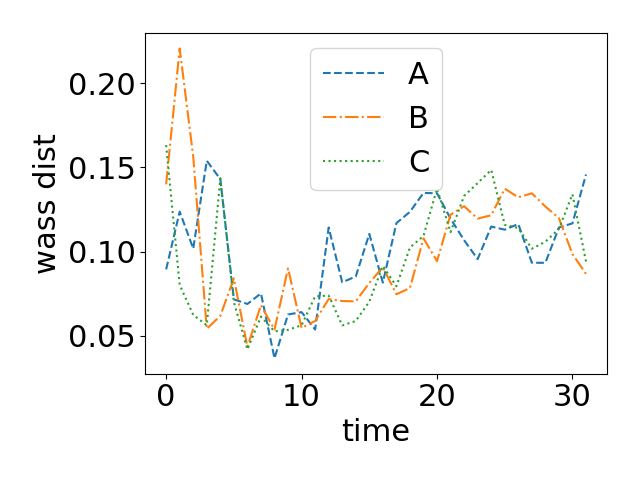}}
        \hfill
    \subfigure[Toggle Switch]{  
    \includegraphics[scale=0.235]{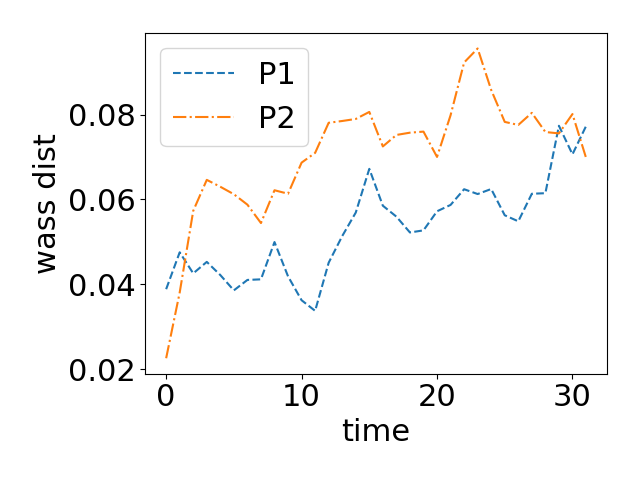}}
        \hfill
    \subfigure[MAPK]{  
    \includegraphics[scale=0.235]{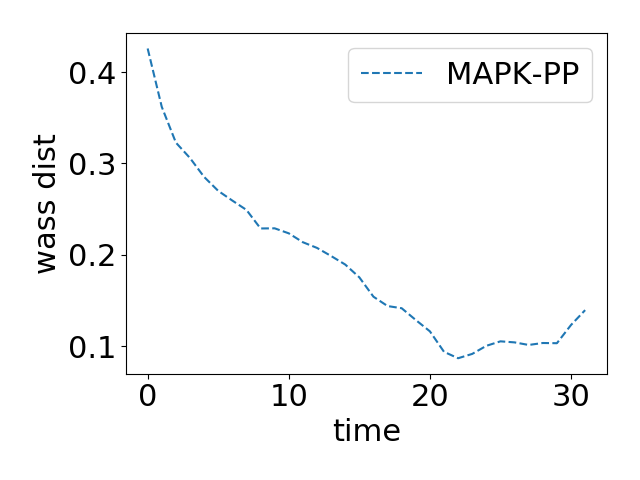}}

\vspace{-1\baselineskip}

\caption{Plots of the error over time for each model and each species. Errors are computed using the Wasserstein distance over the entire test set. Generated trajectories have been keep scaled to the interval $[-1,1]$ so that the scale of the system does not affect the scale of the error measure.\vspace{-0.5cm}}\label{fig:wass_errors}
\end{figure}

\paragraph{Measuring error propagation.}
The reconstruction accuracy of the proposed abstraction procedure is performed on test sets consisting of $25$ different initial settings. For each of these points $2$K SSA trajectories represent the empirical approximation of the true distribution over $S^H$. 
From each of these initial settings we also simulate $2$K abstract trajectories. Given a species $i\in\{1,\dots n\}$ and a time step $j\in\{1,\dots H\}$, we have the real one-dimensional distribution $\eta_{i,j}$ and the generated abstract distribution $\hat{\eta}_{i,j}$, where $\eta_{i,j}$ denotes the counts of species $i$ at time $t_j$ in a trajectory $\eta_{[1,H]}$. In order to quantify the reconstruction error, we compute five quantities: the Wasserstein distance among the two one-dimensional distributions, the absolute and relative difference among the two means and the absolute and relative difference among the two variances. By doing so, we are capable of seeing whether the error propagates in time and whether some species are harder to reconstruct than others.
The error plots for the Wasserstein distance
are shown in Figure~\ref{fig:wass_errors}. Plots of means and variances distances are provided in Appendix~\ref{sec:plots} (Fig.~\ref{fig:avg_means_errors}-~\ref{fig:avg_vars_rel_errors}). In addition, for two-dimensional models, i.e. eSIRS, Toggle Switch and MAPK, we show the landscapes of these five measures of the reconstruction error at three different time steps: step $t_1$, step $t_{H/2}$ and step $t_H$ (Fig.~\ref{fig:eSIRS_distance_landscapes}-\ref{fig:mapk_distance_landscapes} in Appendix~\ref{sec:plots}).  We observe that, in all the models, each species seems to contribute equally to the global error and, in general, the error stays constant w.r.t. time, i.e., it does not propagate. This was a major concern in previous methods, based on the abstraction of transition kernels. 
In fact, in order to simulate a trajectory of length $H$ the abstract kernel has to be applied iteratively $H$ times. As a consequence, this results in a propagation of the error introduced in the approximation of the transition kernel.

\begin{figure}[ht]
    \centering
    
    \includegraphics[scale=0.25]{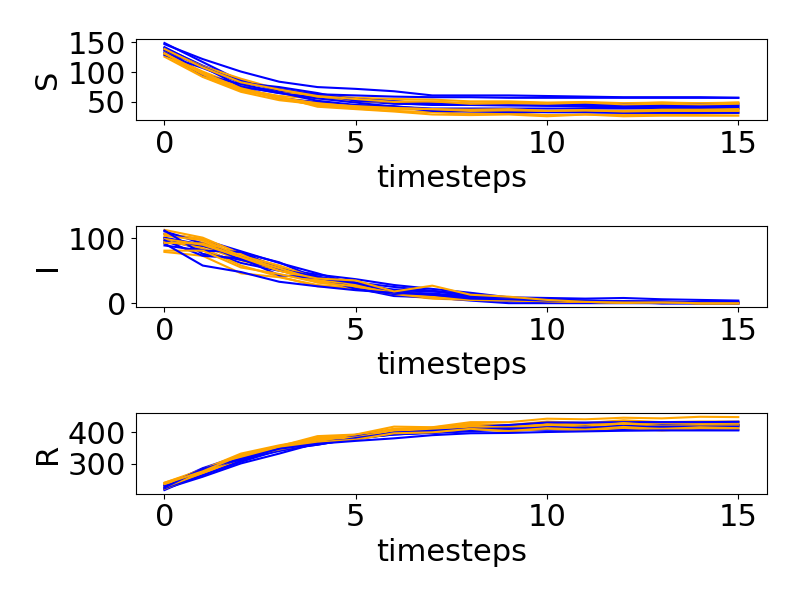}
    \includegraphics[scale=0.25]{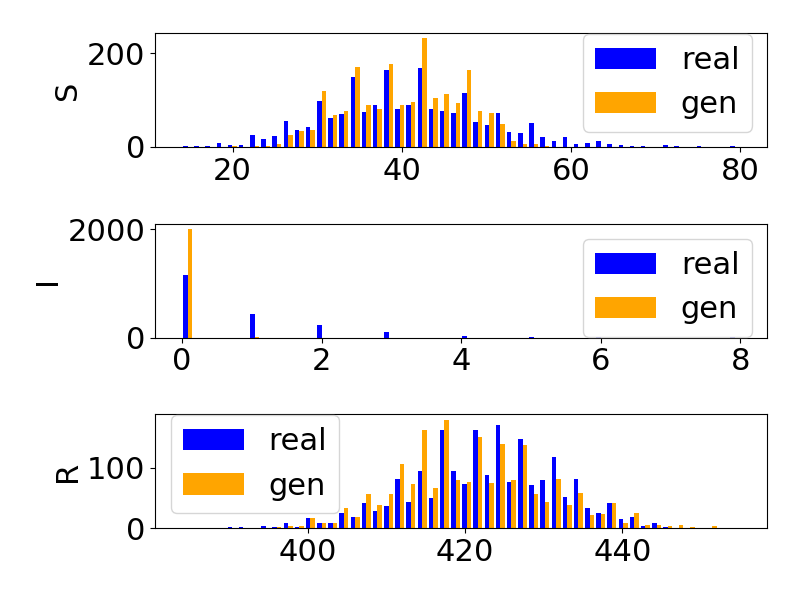}
    
    \vspace{-1\baselineskip}
    
    \caption{SIR
    model: \textbf{(left)} comparison of trajectories generated with a cWCGAN-GP (orange) and the trajectories generated with the SSA algorithm (blue); \textbf{(right)} comparison of the real and generated histogram at the last timestep. Performance on a randomly chosen test point represented by three trajectories: the top one (species S), the central one (species I) and the bottom one (species R).\vspace{-0.5cm}}
    \label{fig:sir_trajectories}
\end{figure}

\begin{figure}[ht]
    \centering
    \includegraphics[scale=0.25]{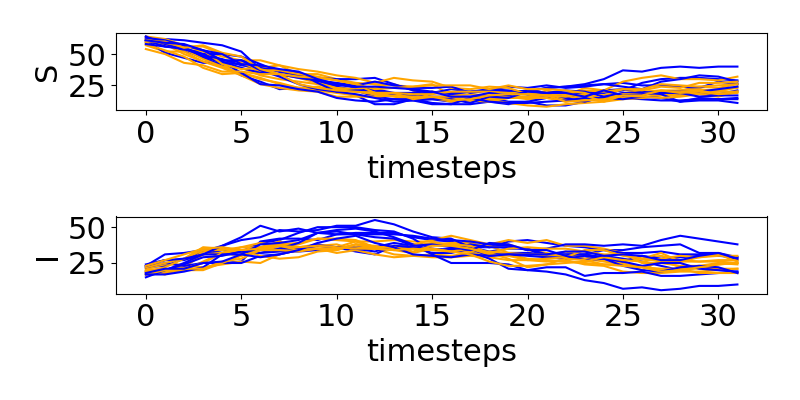}
    \includegraphics[scale=0.25]{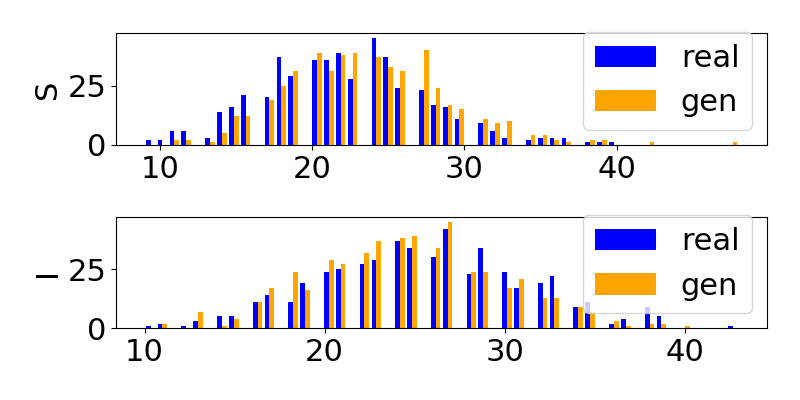}
    \includegraphics[scale=0.25]{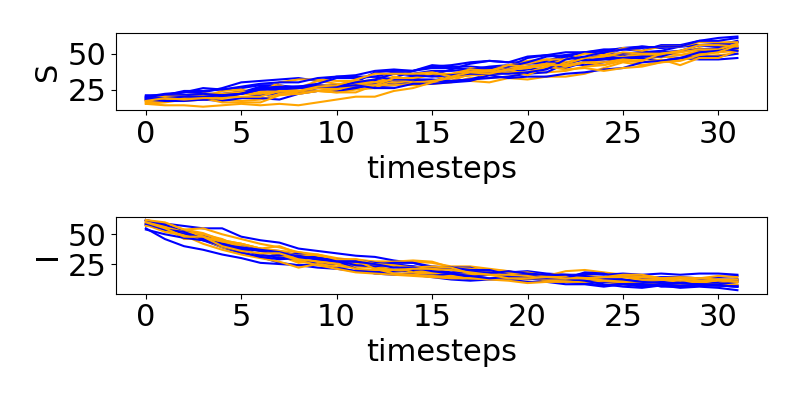}
    \includegraphics[scale=0.25]{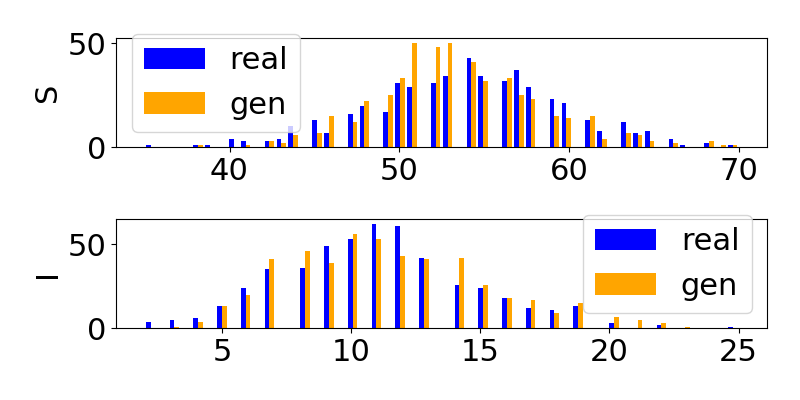}
    
    \vspace{-1\baselineskip}
    
    \caption{e-SIRS model with one varying parameter: \textbf{(left)} comparison of trajectories generated with a cWCGAN-GP (orange) and the trajectories generated with the SSA algorithm (blue); \textbf{(right)} comparison of the real and generated histogram at the last timestep. }
    \label{fig:esir_trajectories_one_par}

    \includegraphics[scale=0.25]{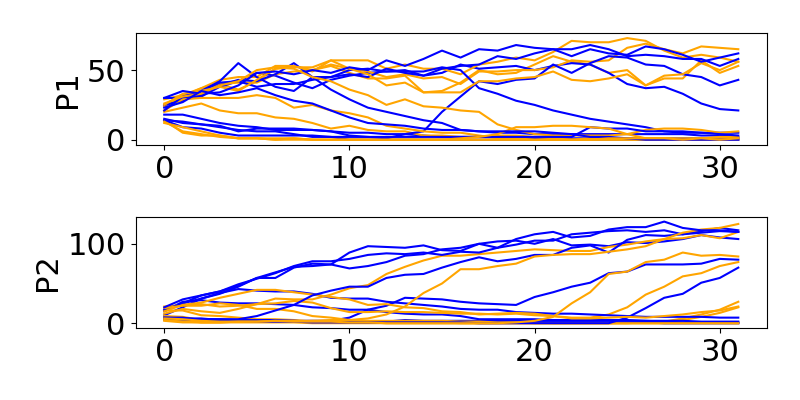}
    \includegraphics[scale = 0.25]{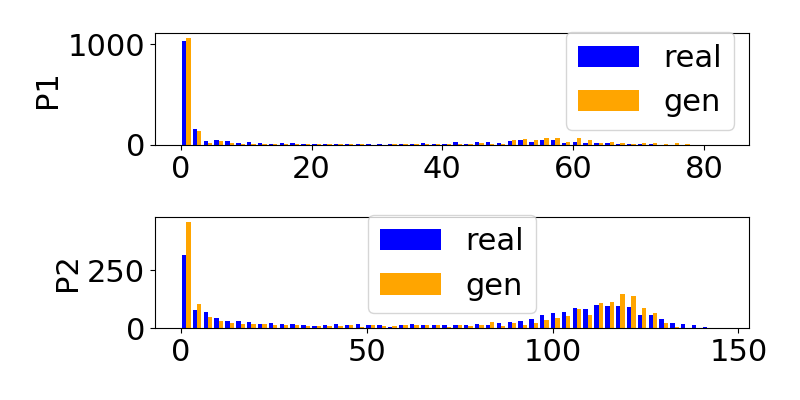}
    
    \vspace{-1\baselineskip}
    
    \caption{Toggle Switch model: \textbf{(left)} comparison of trajectories generated with a cWCGAN-GP (orange) and the trajectories generated with the SSA algorithm (blue); \textbf{(right)} comparison of the real and generated histogram at the last timestep. Performance for a randomly chosen test point represented by a pair of trajectories: the top one (species P1) and the bottom one (species P2).\vspace{-0.5cm}}
    \label{fig:ts_trajectories}
\end{figure}

\begin{figure}[ht]
   \centering
   \includegraphics[scale=0.25]{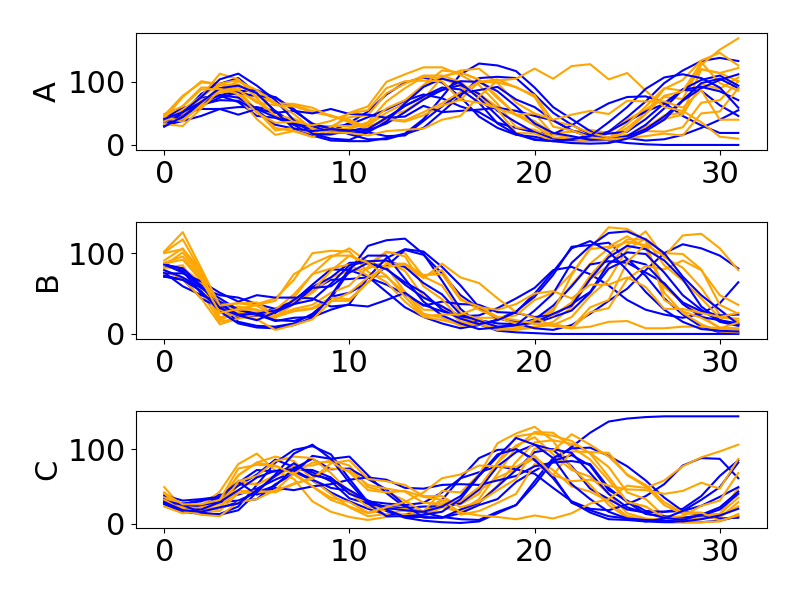}
   \includegraphics[scale=0.25]{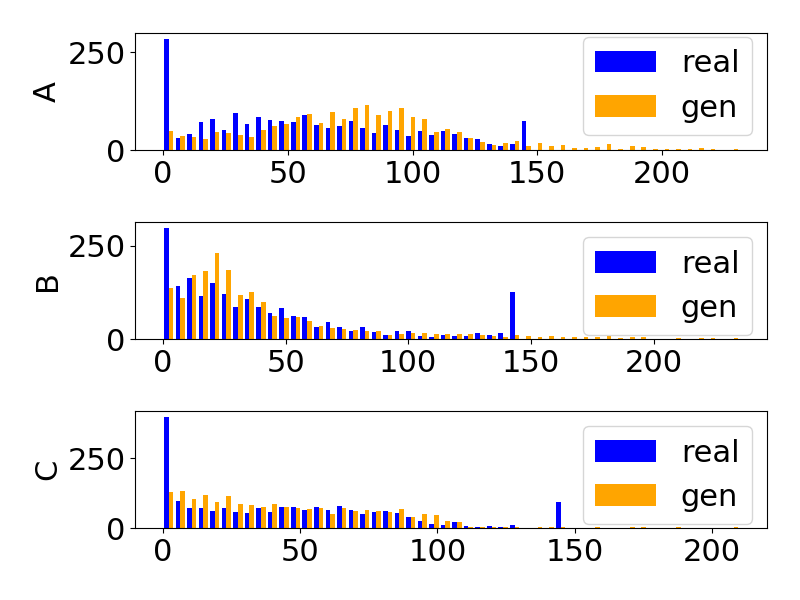}
   
   \vspace{-1\baselineskip}
   
    \caption{Oscilator model: \textbf{(left)} comparison of trajectories generated with a cWCGAN-GP (orange) and the trajectories generated with the SSA algorithm (blue);\textbf{(right)} comparison of the real and generated histogram at the last timestep. Performance on a randomly chosen test point represented by three trajectories: the top one (species A), the central one (species B) and the bottom one (species C). \vspace{-0.5cm}
   }\label{fig:clock_trajectories}
\end{figure}

\begin{figure}[ht]
   \centering
   \includegraphics[scale=0.19]{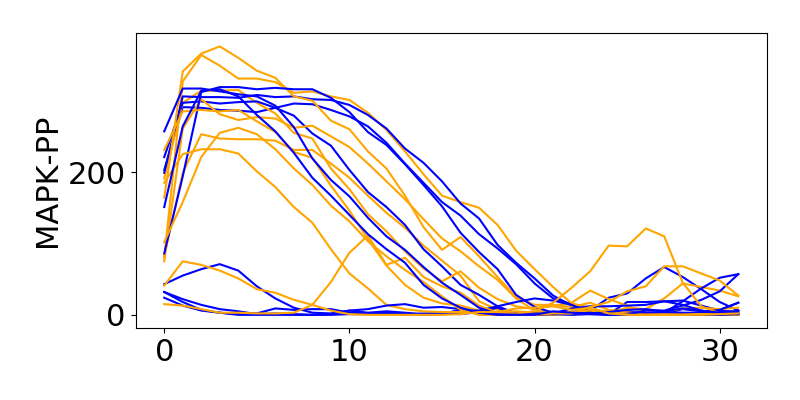}
   \includegraphics[scale=0.19]{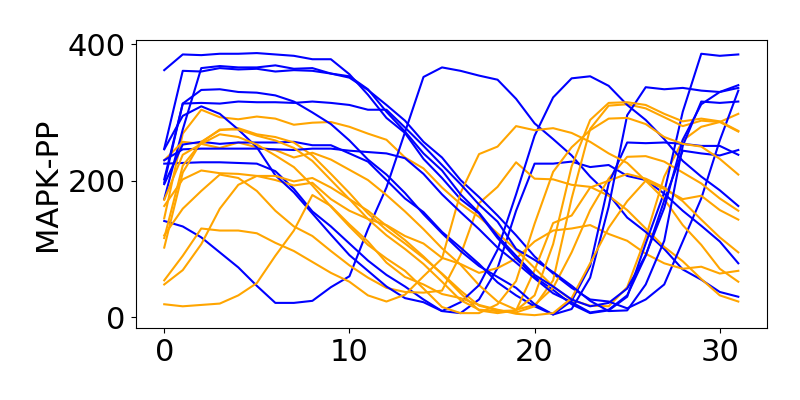}
   \includegraphics[scale=0.19]{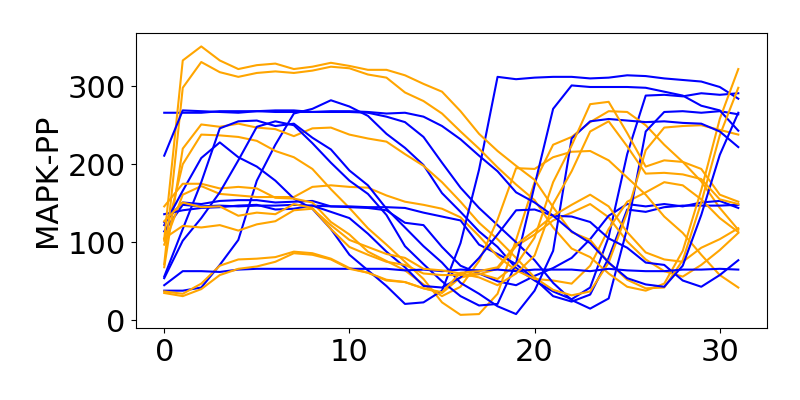}
   \includegraphics[scale=0.19]{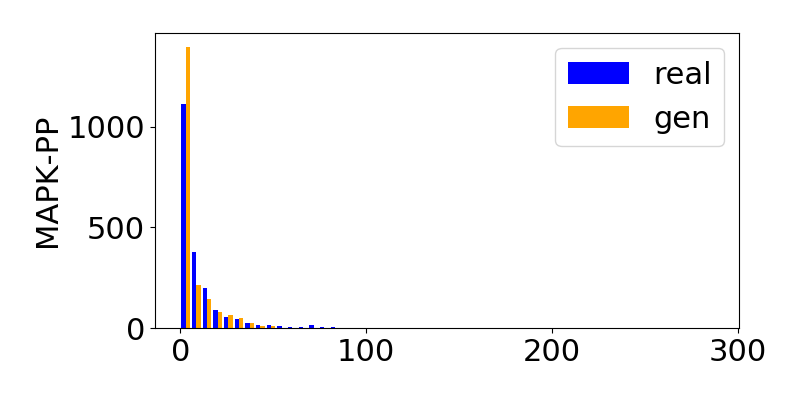}
   \includegraphics[scale=0.19]{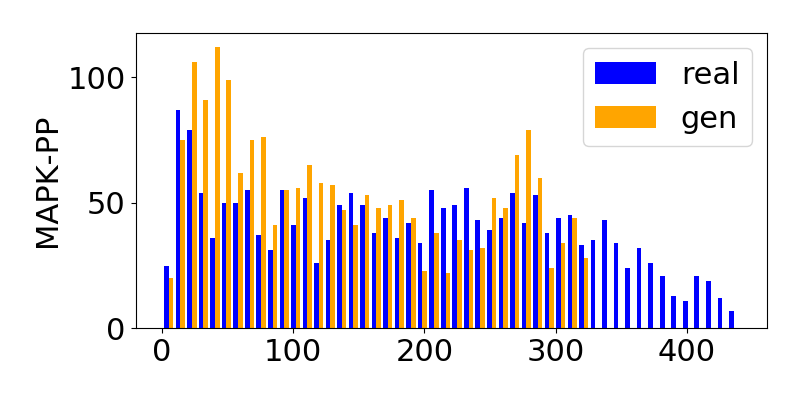}
   \includegraphics[scale=0.19]{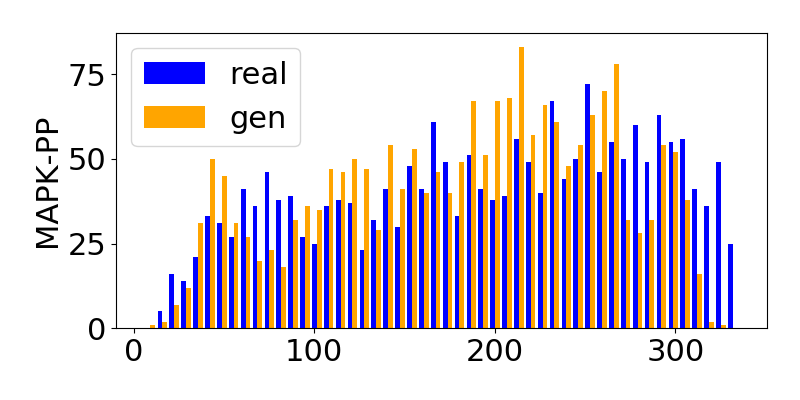}
   
   \vspace{-1\baselineskip}
   
    \caption{MAPK model: \textbf{(top)} comparison of trajectories generated with a cWCGAN-GP (orange) and the trajectories generated with the SSA algorithm (blue);\textbf{(bottom)} comparison of the real and generated histogram at the last timestep. Performance on three, randomly chosen, test points. Each point is represented by the ouput species MAPK\_PP. \vspace{-0.5cm}
   }\label{fig:mapk_trajectories}
\end{figure}

\textbf{SIR.} The results for the SIR model are presented in Fig.~\ref{fig:sir_trajectories} and Fig.~\ref{fig:sir_trajectories_extra} (Appendix~\ref{sec:plots}), which shows the performance on two, randomly chosen, test points. Each point is represented by three trajectories, the top one is for species S, the central one is for species I and the bottom one is for species R. The population size, given by $S+I+R$, is variable. The abstraction was trained on a dataset with fixed parameters, $\theta = \{3,1\}$. Likewise, in the test set only the initial states are allowed to vary.
We observe that our abstraction method is able to capture the absorbing nature of SIR trajectories. It is indeed very important that once state $I=0$ or state $R=N$ are reached, the system should not escape from it. Abstract trajectories satisfy such property without requiring the imposition of any additional constraint.
The empirical distributions, real and generated, at time $t_H$ are almost indistinguishable.

\textbf{e-SIRS.} The e-SIRS model represents our baseline. We train two abstractions: in the first case the model is trained on a dataset with fixed parameters, $\theta=\{2.36, 1.67,  0.9, 0.64\}$, and in the second case we let parameter $\theta_1$ vary as well. Results are very accurate in both scenarios.
In the fixed-parameters case, Fig.~\ref{fig:esir_trajectories} (Appendix~\ref{sec:plots}), the results are shown for two, randomly chosen, initial states. In the second case, Fig.~\ref{fig:esir_trajectories_one_par}, the results are shown on two, randomly chosen, pairs $(s_0,\theta_1)$. Each point is represented by a pair of trajectories, the top one is for species S and the bottom one is for species I. We performed a further analysis on the generalization capabilities of the abstraction learned on the dataset with one varying parameter, using larger test sets and computing mean and standard deviation of the distribution of Wasserstein distances over such sets. The mean stays around $0.04$ with a tight standard deviation ranges from $0.01$ to $0.05$, showing little impact of the chosen conditional setting (see Fig.~\ref{fig:esirs_1p_analysis} in Appendix~\ref{sec:plots}).

\textbf{Toggle Switch.} The results for the Toggle Switch model, on two, randomly chosen, test points, are shown in Fig.~\ref{fig:ts_trajectories}) and Fig.~\ref{fig:ts_trajectories_extra} (Appendix~\ref{sec:plots}). The abstraction was trained on a dataset with fixed symmetric parameters ($kp_i=1,kb_i=1,ku_i=1, kd_i=0.01$ for $i = 1,2$). Likewise, in the test set only the initial states are allowed to vary. In this model, we tried to abstract only trajectories of the proteins $P1$ and $P2$, which are typically the observable species, ignoring the state of the genes. By doing so, we reduce the dimensionality of the problem but we also lose some information about the full state of the system. Nonetheless, the cWCGAN-GP abstraction is capable of capturing the bistable behaviour of such trajectories. In Fig.~\ref{fig:ts_trajectories}, each point is represented by two trajectories, the top one is for species $P1$, whereas the bottom one is for species $P2$.

\textbf{Oscillator.} The results for the Oscillator model, on two, randomly chosen, test points, are shown in Fig.~\ref{fig:clock_trajectories} and Fig.~\ref{fig:clock_trajectories_extra} (Appendix~\ref{sec:plots}). The abstraction was trained on a dataset with fixed parameter ($\theta = 1$). Likewise, in the test set only the initial states are allowed to vary. Each point is represented by three trajectories, the top one is for species $A$, the central one is for species $B$ and the bottom one is for species $C$. The abstract trajectories well capture the oscillating behaviour of the system. 

\textbf{MAPK.} The results for the MAPK model, on three, randomly chosen, test points, are shown in Fig.~\ref{fig:mapk_trajectories}. The abstraction was trained on a dataset considering only a varying $V_1$ parameter and the dynamics of species $MAPK\_PP$. 
This case study represents a complex scenario in which the abstract distribution should capture the marginalization over the other seven unobserved variables. Moreover, the emergent behaviour of the only observed variable, $MAPK\_PP$, is strongly influenced by the input parameter $V_1$ and further  amplified by the multi-scale nature of the cascade: for some values of $V_1$ the system oscillates, whereas for others it stabilizes around an equilibrium. Results show that our abstraction technique is flexible enough to capture such sensitivity.

\vspace{-0.2cm}

\subsection{Discussion}
Previous approaches to model abstraction, see Related work in Section~\ref{sec:introduction}, focus on approximating the transition kernel, meaning the distribution of possible next states after a time $\Delta t$, rather than learning the distribution of full trajectories of length $H$. The main reason for such choice is the limited scalability of the tool used for learning the abstraction. In fact, learning a distribution over $S^H\subseteq\mathbb{N}^{H\times n}$ with a Mixture Density Network is unfeasible even for small $H$. Moreover, in learning to approximate the transition kernel one must split the SSA trajectories of the dataset in pairs of subsequent states. By doing so, a lot of information about the temporal correlation among states is lost. Having a tool strong and stable enough to learn distributions over $S^H$ allows us to preserve this information and make abstraction possible even for systems with a complex dynamics, which the abstraction of the transition kernel was failing to capture. For instance, we are now able to abstract the transient behaviour of multi-stable or oscillating systems. When attempting to abstract the transition kernel, either via MDN or via c-GAN, for such complex systems, we did not succeed in learning 
meaningful solutions.
A collateral advantage in generating full trajectories, rather than single subsequent states, is that it introduces an additional computational speed-up in the time required to generate a large pool of trajectories of length $H$. For instance, if a cWGAN is used to approximate the transition kernel, it takes around $31$ seconds to simulate the $50$K trajectories of length $32$ present in the test set. Our trajectory-based method takes only $3.4$ seconds to generate the same number of trajectories. Furthermore, our cWCGAN-GP was trained with relative small datasets, which leaves room for further improvements where needed.
An additional strength of our method is that one can train the abstract model only on species that are observable, reducing the complexity of the CRN model while preserving an accurate reconstruction for the species of interest. Once again, this was not possible with transition kernels and it may be extremely useful in real world applications.

In general, the cWCGAN-GP approximation does not provide any statistical guarantee about the reconstruction error. In addition, the set of observations used to learn the abstraction is rather small, typically $10$ samples for each initial setting. Therefore, it is not surprising that the real and the abstract distributions are not indistinguishable from a statistical point of view, as shown in Appendix~\ref{sec:stat_test}.
However, the abstract model is actually capable of capturing, from the little amount of information provided, the emergent features of the behaviour of the original system, such as multimodality or oscillations. 
In this regard, formal languages can be used to formalize and check such qualitative properties. In particular, we can check whether the satisfaction probability (of non rare events) is similar in real and abstract trajectories. Examples are shown in Appendix~\ref{sec:satisf}. Furthermore, such quantification of qualitative properties can be used to measure how good the reconstruction is. As future work, we intend to use it as query strategy for an active learning approach, so that the obtained abstract model is driven in the desired direction.

\vspace{-0.25cm}

\section{Conclusions}\label{sec:conclusions}
In the paper we presented a technique to abstract the simulation process of stochastic trajectories for various CRNs. The WGAN-based abstraction improves considerably the computational efficiency, which is no more related to the complexity of the underlying CRN. 
This would be extremely helpful in all those applications in which a large number of simulations is required, i.e., applications whose solution is unfeasible via SSA simulation. It would enable the simulation of multi-scale models for very large populations, it would speed-up statistical model checking \cite{younes2006statistical} and it can be used in particular cases of parameter estimation, for example when only few parameters have to be estimated multiple times.
In conclusion, the c-WCGAN-based solution to model abstraction perform well in scenarios that are very complex and challenging, requiring relatively little data and very little fine-tuning.

As future work, we plan to study how our abstraction technique works on real data. In this regard, we do not aim at capturing the underlying dynamical system, but we would rather be able to reproduce the trajectories observed in real applications. 
A great strength of our method, compared to state of the art solutions, is that it is able to generate trajectories only for a subset of the species present in the system domain, ignoring the information that is not observable, even during the training phase.
Another interesting extension is to adapt our technique to sample bridging trajectories, where both the initial and the terminal states are fixed. Typically, the simulation of such trajectories requires expensive Monte Carlo simulations, which makes clear the benefits of resorting to model abstraction.

{\footnotesize	 \noindent\textbf{Acknowledgements}
This work has been partially supported by 
the Italian PRIN project ``SEDUCE'' n.\ 2017TWRCNB.}

\bibliographystyle{splncs04}
\bibliography{biblio}

\appendix

\section{Case Studies}\label{sec:casestudies}
\begin{itemize}
\item \textbf{SIR Model (Absorbing state).} 
The SIR epidemiological model describes a population divided in three mutually exclusive groups: susceptible (S), infected (I) and recovered (R). The system state at time $t$ is $\eta_t= (S_t, I_t, R_t)$. The possible reactions, given by the interaction of individuals (representing the molecules of a CRN), are the following:
\begin{itemize}
    \item $R_1: S+I\xrightarrow{\theta_1\cdot I_tS_t/(S_t+I_t+R_t)} 2I$ (infection),
    \item $R_2: I\xrightarrow{\theta_2\cdot I_t} R$  (recovery).
\end{itemize}
The model describes the spread, in a population, of an infectious disease that grants immunity to those who recover from it. As the SIR model is well-known and stable, we use it as a testing ground for our GAN-based abstraction procedure. The ranges for the initial state are $S_0, I_0, R_0 \in [30, 200]$.
An important aspect of the SIR model is the presence of an absorbing states. In fact, when $I = 0$ or when $R=N$ no more reaction can take place. 

\item \textbf{Ergodic SIRS Model.} Small perturbations of the SIR model force the system to be ergodic. We called this revised version ergodic SIRS (eSIRS). This model has no absorbing state. In particular, we assume that the population is not perfectly isolated, meaning there is always a chance of getting infected from some external individuals. In addition, we also assume that immunity is only temporary.
The possible reactions are now the following:
\begin{itemize}
    \item $R_1: S+I\xrightarrow{\theta_1\cdot I_tS_t/(S_t+I_t+R_t)+\theta_2\cdot S_t} 2I$ (infection),
    \item $R_2: I\xrightarrow{\theta_3\cdot I_t} R$  (recovery),
    \item $R_3: R\xrightarrow{\theta_4\cdot R_t} S$  (immunity loss),
\end{itemize}
Both epidemiological models are essentially unimodal.
The ranges for the initial state are $S_0, I_0, R_0 \in [0, N]$ such that $S_0 + I_0 + R_0 = N$. In our experiments $N= 100$. The range for parameter $\theta_1$ is $[0.5,5]$.

\item \textbf{Genetic Toggle Switch Model (Bistability).} 
The toggle switch is a well-known bistable biological circuit. Briefly, this system consists of two genes, $G_1$ and $G_2$, that mutually repress each other. The system displays two stable equilibrium states in which either of the two gene products represses the expression of the other gene. 
The possible reactions are:
\begin{itemize}
    \item $prod_i: G_i^{on}\xrightarrow{kp_i\cdot G_i^{on}} G_i^{on}+P_i$, for $i=1,2$;,
    \item $bind_i: 2P_j+G_i^{on}\xrightarrow{kb_i\cdot G_i^{on}\cdot P_j\cdot(P_j-1)} G_i^{off}$, for $i=1,2$ and $j=2,1$ resp.;
    \item $unbind_i: G_i^{off}\xrightarrow{ku_i\cdot G_i^{off}} G_i^{on}+2P_j$, for $i=1,2$ and $j=2,1$ resp.;
    \item $deg_i: P_i\xrightarrow{kd_i\cdot Pi} \emptyset$, for $i = 1,2$.
\end{itemize}
The ranges for the initial state are $G_{1,0}, G^{on}_{2,0}\in\{0,1\}$ and  $P_{1,0}, P_{2,0} \in [5, 20]$.

\item \textbf{Oscillator Model.} 
The oscillator circuit consists of three species A, B and C and three reactions, in which A converts B to itself, B converts C to itself, and C converts A to itself.
The three species regulate each other in a cyclic manner. This circuit was found to exhibit oscillations in the concentrations of the three species. 
\begin{itemize}
    \item $R_1: A+B\xrightarrow{{\tiny \theta\cdot\frac{A\cdot B}{A+B+C}}} 2A$ (B transformation),
    \item $R_2: B+C\xrightarrow{{\tiny \theta\cdot\frac{B\cdot C}{A+B+C}}} 2B$ (C transformation),
    \item $R_3: C+A\xrightarrow{{\tiny \theta\cdot\frac{C\cdot A}{A+B+C}}} 2C$ (A transformation).
\end{itemize}  
The ranges for the initial state are $A_0, B_0, C_0 \in [20, 100]$.

\item \textbf{MAPK Model.}

Mitogen-activated protein kinase cascade is a particular type of signal transduction into protein phosphorylation (PP) whose function is the amplification of a signal. The sensitivity increases with the number of cascade levels, such that a small change in a stimulus results in a large change in the response.
Negative feedback from MAPK-PP to the MAKKK activating reaction with ultra-sensitivity to a input stimulus, governed by parameter $V_1$.

\begin{itemize}
\item $\it R_1: MKKK \xrightarrow{{\tiny V_1\cdot MKKK/( (1+(MAPK\_PP/K_l)^n)\cdot (K_1+MKKK) )}} MKKK\_P$,
	
\item $\it R_2:  MKKK\_P \xrightarrow{{\tiny V_2\cdot MKKK\_P/(K_2+MKKK\_P)}} MKKK $,
	
\item $\it R_3:	MKK \xrightarrow{{\tiny k_3\cdot MKKK\_P\cdot MKK/(K_3+MKK)}} MKK\_P $,
	
\item $\it R_4:  MKK\_P  \xrightarrow{{\tiny k_4\cdot MKKK\_P\cdot MKK\_P/(K_4+MKK\_P)  }}     MKK\_PP$,
	
\item $\it R_5: 	MKK\_PP
\xrightarrow{{\tiny V_5\cdot MKK\_PP/(K_5+MKK\_PP) }} 
MKK\_P $,
	
\item $\it R_6:
	MKK\_P 
	\xrightarrow{{\tiny V_6\cdot MKK\_P/(K_6+MKK\_P)}}
	MKK$,
	
\item $\it R_7:
	MAPK 
	\xrightarrow{{\tiny k_7\cdot MKK\_PP\cdot MAPK/(K_7+MAPK)}}
   MAPK-P$,
\item $\it R_8:
	MAPK\_P 
	\xrightarrow{{\tiny k_8\cdot MKK-PP\cdot MAPK\_P/(K_8+MAPK\_P)}}
    MAPK\_PP$,
    
\item $\it R_9:
	MAPK\_PP 
	\xrightarrow{{\tiny V_9\cdot MAPK\_PP/(K_9+MAPK\_PP)}}
    MAPK\_P$,
    
\item $\it R_{10}:
	MAPK\_P 
	\xrightarrow{{\tiny V_{10}\cdot MAPK\_P/(K_{10}+MAPK\_P)}}
	MAPK$.

\end{itemize}

\end{itemize}
The ranges for the initial state are: $MKKK_0, MKKK_0\_P \in [0, 100]$ such that $MKKK_0 + MKKK_0\_P = 100$; $MKK_0, MKK_0\_P, MKK_0\_PP \in [0, 300]$ such that $MKK_0 + MKK_0\_P+ MKK_0\_PP = 300$; $MAPK_0, MAPK_0\_P, MAPK_0\_PP \in [0, 300]$ such that $MAPK_0 + MAPK_0\_P+ MAPK_0\_PP = 300$.
\section{Additional Plots}
\label{sec:plots}
Here we present the remaining plots showing a qualitative evaluation of the performances of the abstract models.
For each model, we present a small batch of trajectories, both real and abstract (plots on the left column). From the plots of such trajectories we can appreciate if the abstract trajectories are similar to real ones and if they capture the most important macroscopic behaviors. We also show the histograms of the empirical distributions at time $t_H$ for each species (plots on the right column) to quantify the behavior over all the $2$K trajectories present in the test set. In particular, Fig. \ref{fig:sir_trajectories} shows the results for the SIR case study, Fig. \ref{fig:esir_trajectories_one_par} shows the results for the e-SIRS model with one varying parameter, Fig. \ref{fig:ts_trajectories} shows the results for the Toggle Switch model and, finally, Fig. \ref{fig:clock_trajectories} shows the results for the Oscillator model.

\begin{figure}[ht]
    \centering
\includegraphics[scale=0.25]{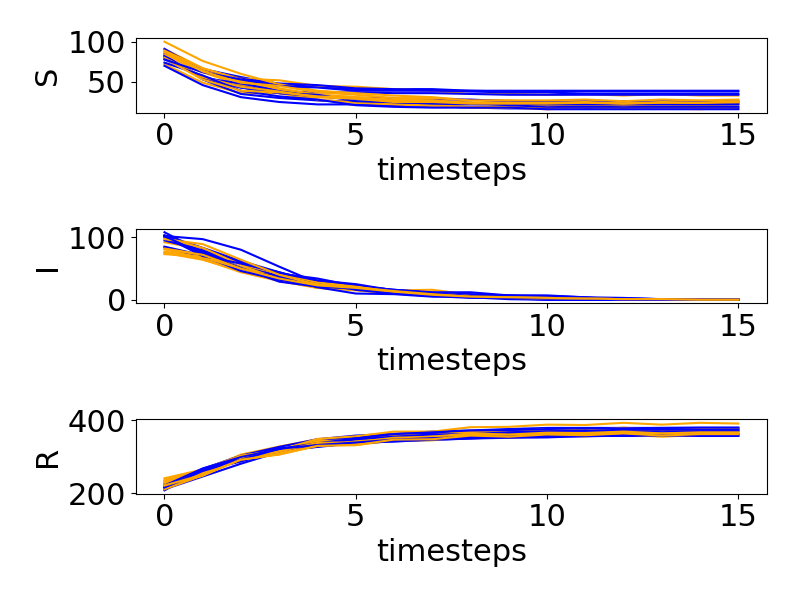}
    \includegraphics[scale=0.25]{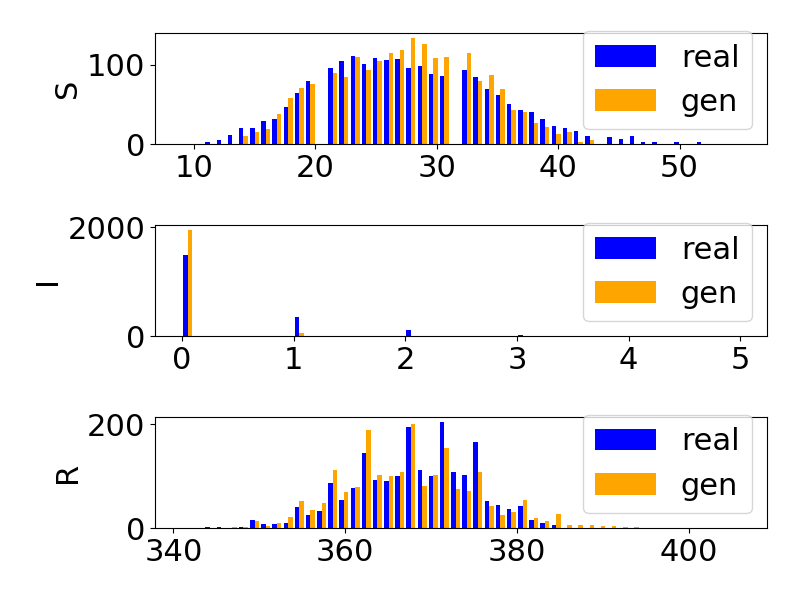}
    
    \caption{SIR
    model: \textbf{(left)} comparison of trajectories generated with a cWCGAN-GP (orange) and the trajectories generated with the SSA algorithm (blue); \textbf{(right)} comparison of the real and generated histogram at the last timestep. Performance on a randomly chosen test point represented by three trajectories: the top one (species S), the central one (species I) and the bottom one (species R).\vspace{-0.5cm}}
    \label{fig:sir_trajectories_extra}
    \end{figure}

\begin{figure}[ht]
    \centering
    \includegraphics[scale=0.25]{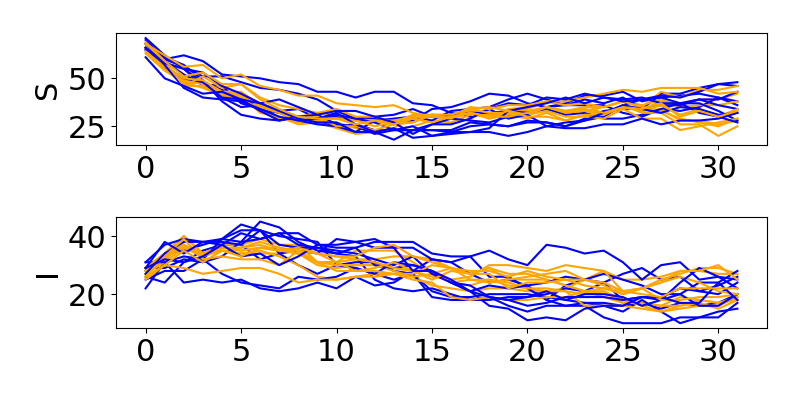}
    \includegraphics[scale=0.25]{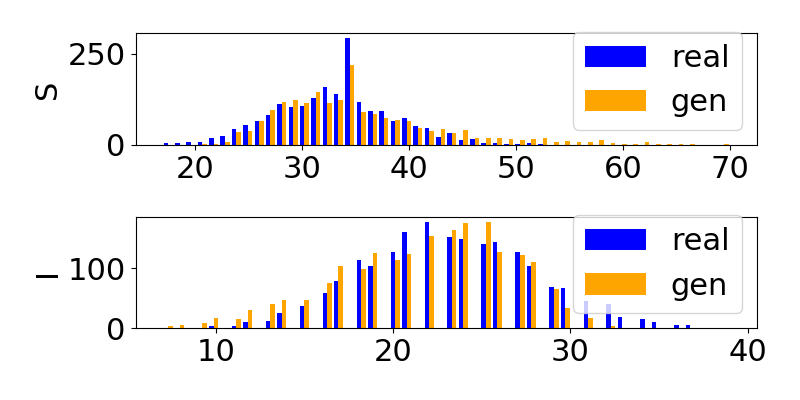}
    \includegraphics[scale=0.25]{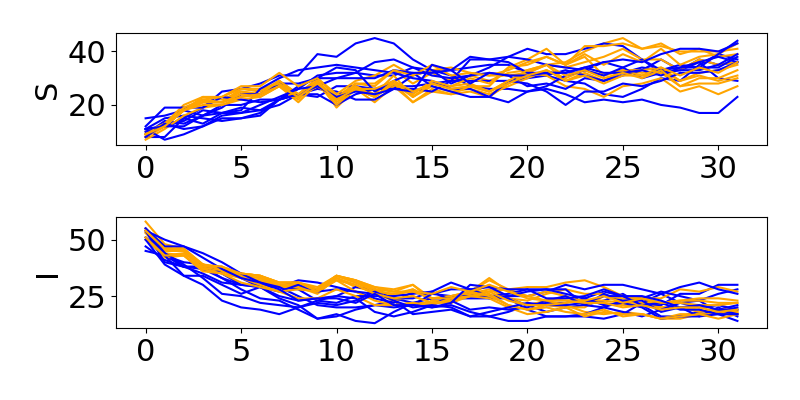}
    \includegraphics[scale=0.25]{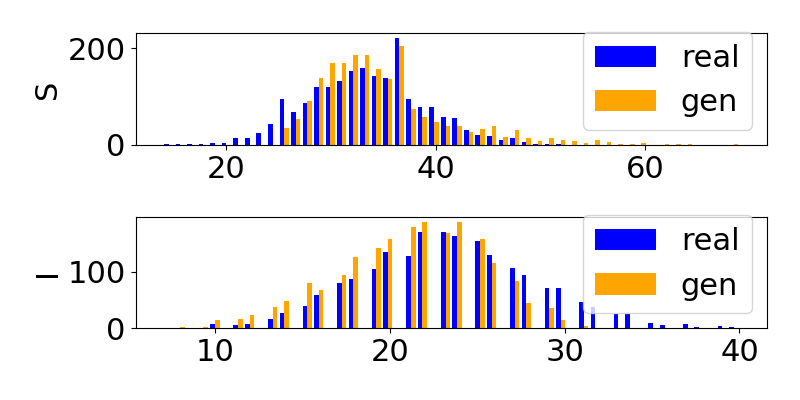}
    \caption{e-SIRS
    model: \textbf{(left)} comparison of trajectories generated with a cWCGAN-GP (orange) and the trajectories generated with the SSA algorithm (blue); \textbf{(right)} comparison of the real and generated histogram at the last timestep. Performance for two, randomly chosen, test points. Each point is represented by a pair of trajectories: the top one (species S) and the bottom one is for  (species I).\vspace{-0.5cm}}
    \label{fig:esir_trajectories}
    \end{figure}

\begin{figure}[ht]
    \centering

    \includegraphics[scale=0.25]{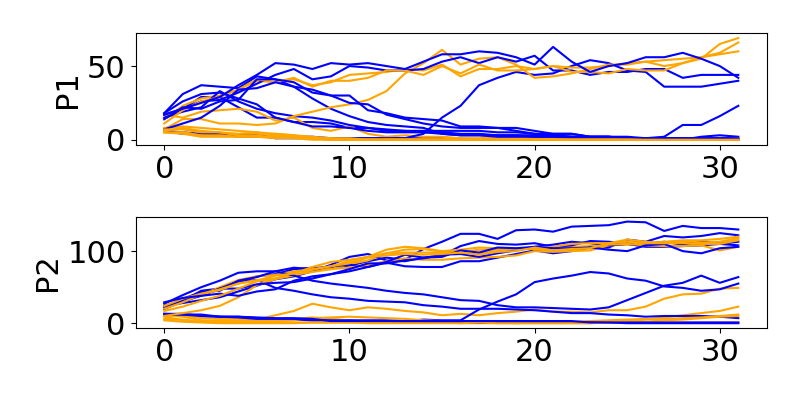}
    \includegraphics[scale = 0.25]{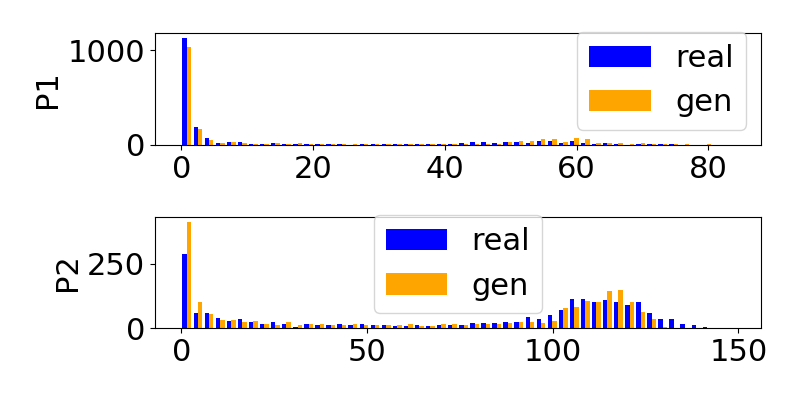}
    
    \caption{Toggle Switch model: \textbf{(left)} comparison of trajectories generated with a cWCGAN-GP (orange) and the trajectories generated with the SSA algorithm (blue); \textbf{(right)} comparison of the real and generated histogram at the last timestep. Performance for a randomly chosen, test point represented by a pair of trajectories: the top one (species P1) and the bottom one (species P2).\vspace{-0.5cm}}
    \label{fig:ts_trajectories_extra}
\end{figure}

\begin{figure}[ht]
   \centering
   
   \includegraphics[scale=0.25]{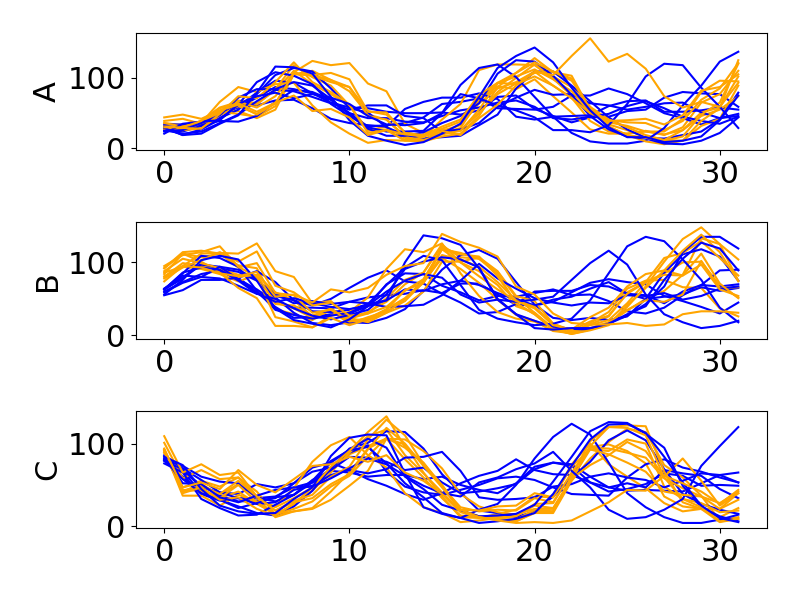}
   \includegraphics[scale=0.25]{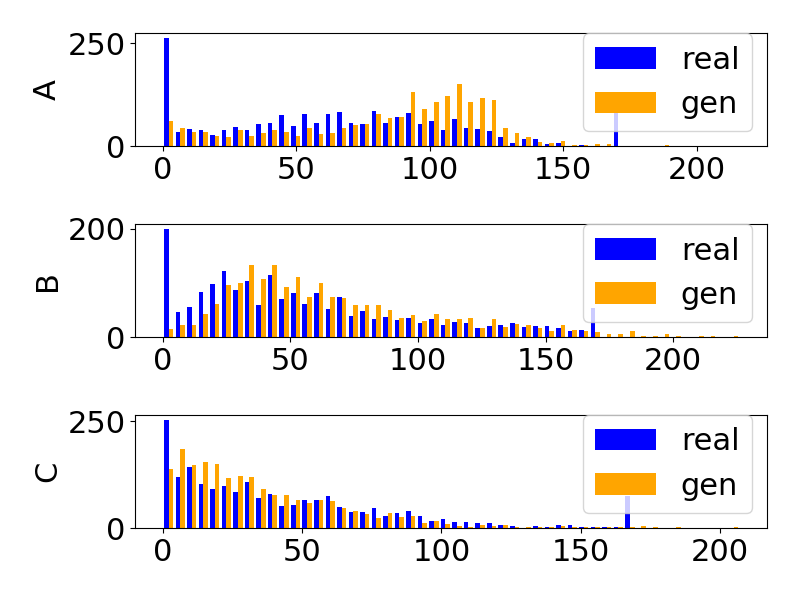}
    \caption{Oscilator model: \textbf{(left)} comparison of trajectories generated with a cWCGAN-GP (orange) and the trajectories generated with the SSA algorithm (blue);\textbf{(right)} comparison of the real and generated histogram at the last timestep. Performance on a randomly chosen test point represented by three trajectories: the top one (species A), the central one (species B) and the bottom one (species C). \vspace{-0.5cm}
   }\label{fig:clock_trajectories_extra}
\end{figure}

\begin{figure}[ht]
    \centering
    
    \subfigure[eSIRS]{  \includegraphics[scale=0.23]{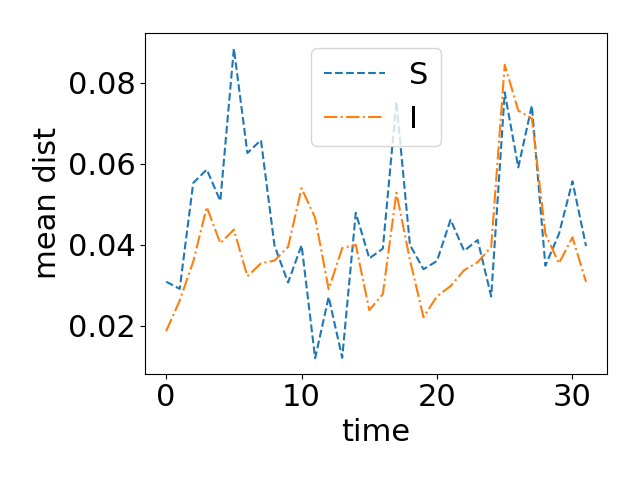}}
    \subfigure[eSIRS-1P]{  
    \includegraphics[scale=0.23]{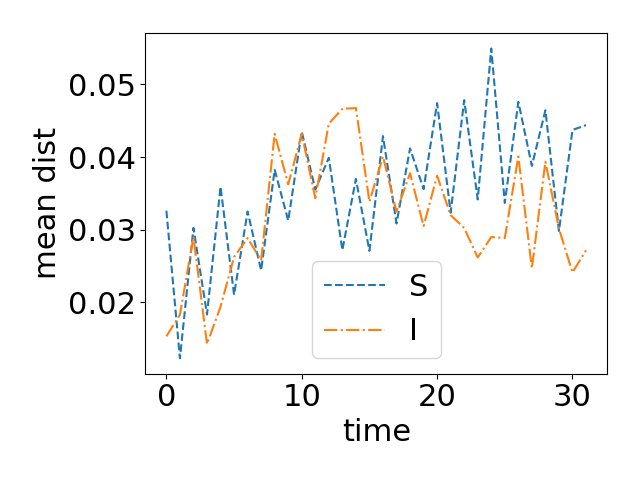}}
    \subfigure[SIR]{  
    \includegraphics[scale=0.23]{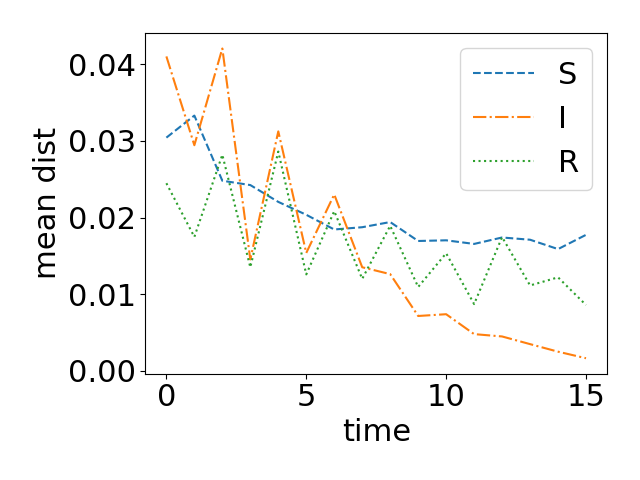}}
    \subfigure[Oscillator]{  
    \includegraphics[scale=0.23]{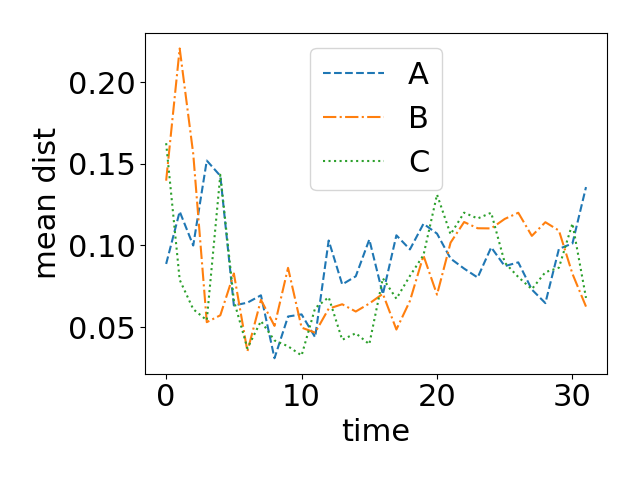}}
    \subfigure[Toggle Switch]{  
    \includegraphics[scale=0.23]{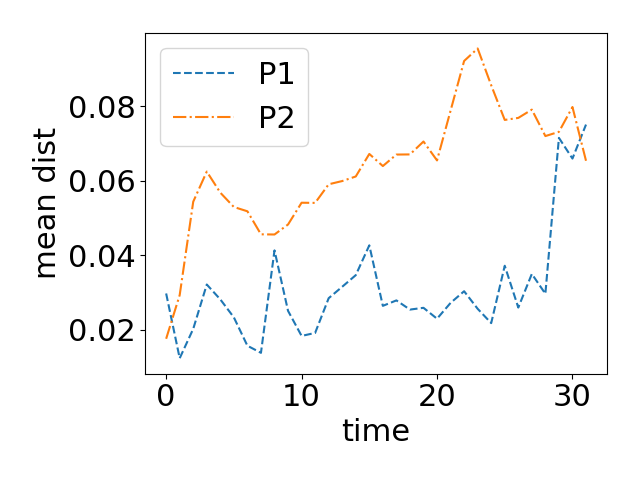}}
     \subfigure[MAPK]{  
    \includegraphics[scale=0.23]{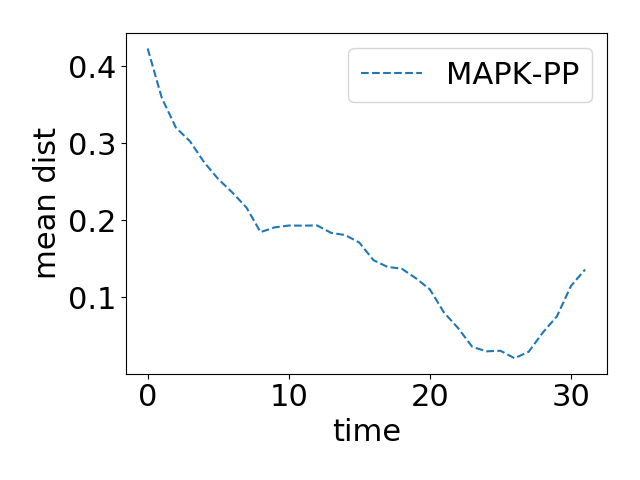}}

    \caption{Plots of the average difference in the means over time for each model and each species. Errors are computed over the entire test set. Generated trajectories have been keep scaled to the interval $[-1,1]$ so that the scale of the system does not affect the scale of the error measure.}\label{fig:avg_means_errors}

    
    \subfigure[eSIRS]{  \includegraphics[scale=0.23]{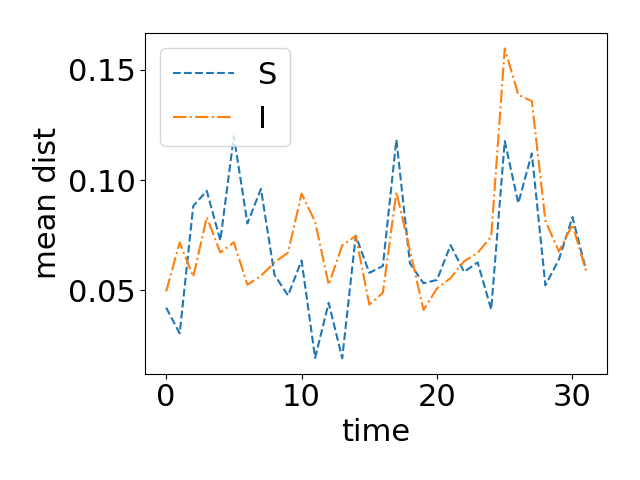}}
    \subfigure[eSIRS-1P]{  
    \includegraphics[scale=0.23]{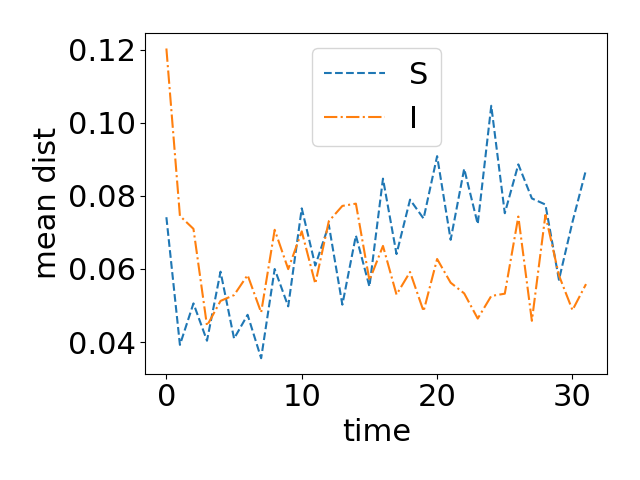}}
    \subfigure[SIR]{  
    \includegraphics[scale=0.23]{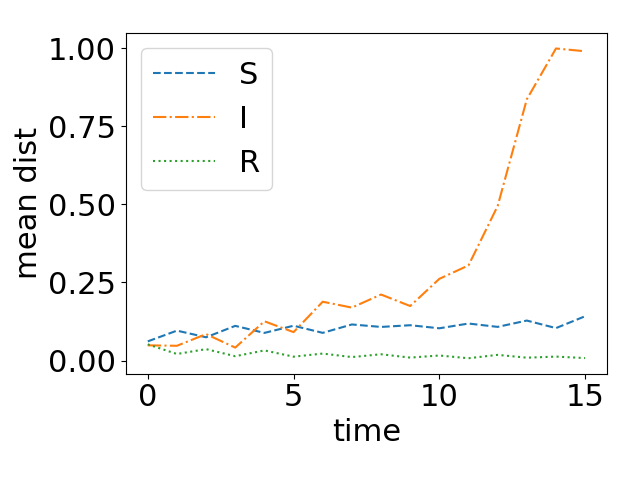}}
    \subfigure[Oscillator]{  
    \includegraphics[scale=0.23]{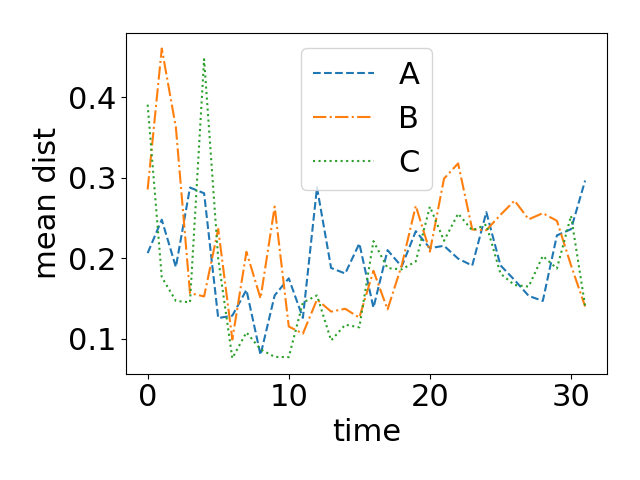}}
    \subfigure[Toggle Switch]{  
    \includegraphics[scale=0.23]{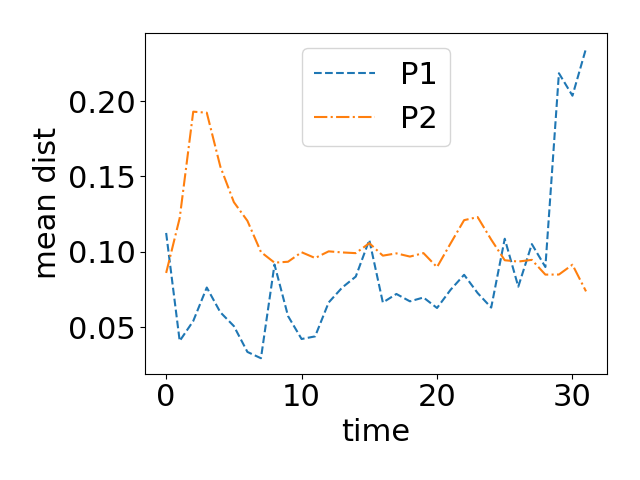}}
     \subfigure[MAPK]{  
    \includegraphics[scale=0.23]{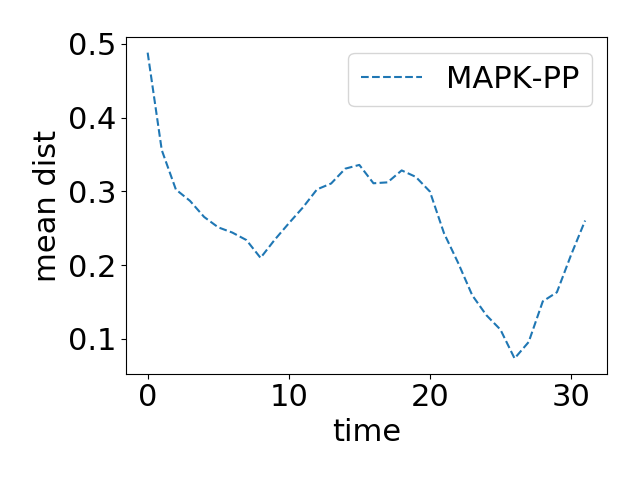}}

    \caption{Plots of the average relative difference in the means over time for each model and each species. Errors are computed over the entire test set. Generated trajectories have been keep scaled back to $\mathbb{N}$.}\label{fig:avg_means_rel_errors}
\end{figure}

\begin{figure}[ht]
    \centering
    \subfigure[eSIRS]{  
    \includegraphics[scale=0.23]{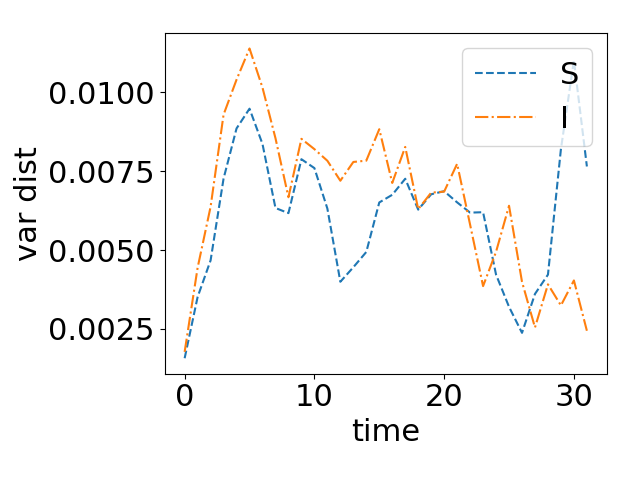}}
    \subfigure[eSIRS-1P]{  
    \includegraphics[scale=0.23]{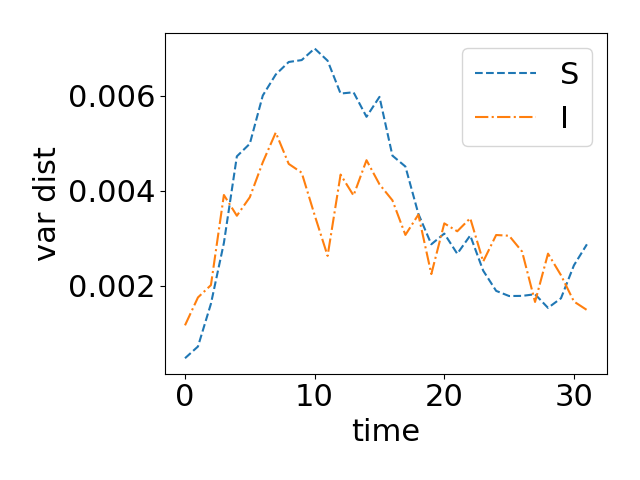}}
    \subfigure[SIR]{  
    \includegraphics[scale=0.23]{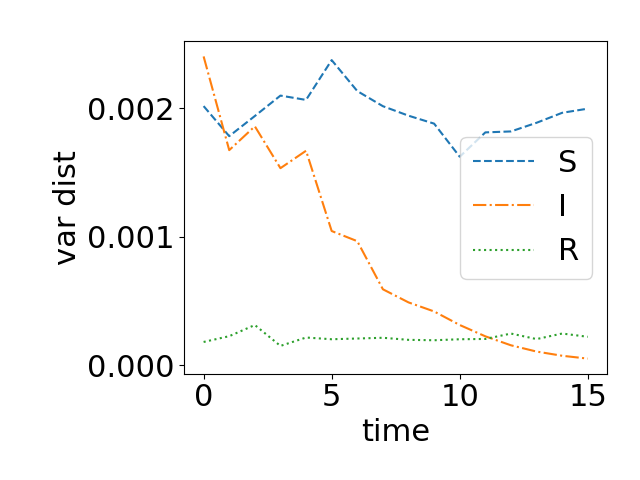}}
    \subfigure[Oscillator]{  
    \includegraphics[scale=0.23]{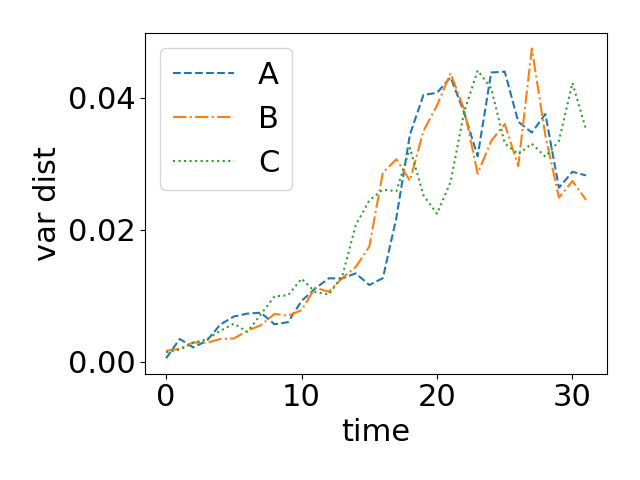}}
    \subfigure[Toggle Switch]{  
    \includegraphics[scale=0.23]{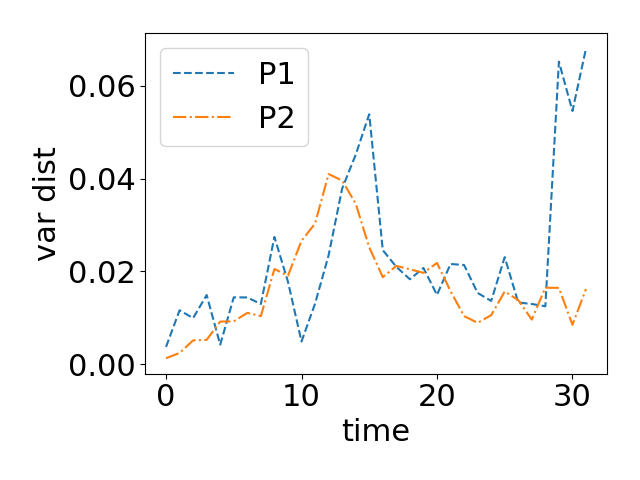}}
    \subfigure[MAPK]{  
    \includegraphics[scale=0.23]{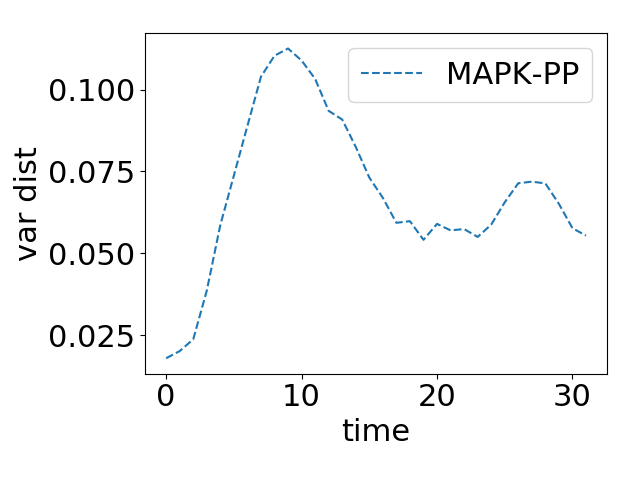}}

    \caption{Plots of the average difference in the variances over time for each model and each species. Errors are computed over the entire test set. Generated trajectories have been keep scaled to the interval $[-1,1]$ so that the scale of the system does not affect the scale of the error measure.}\label{fig:avg_vars_errors}
    
    \subfigure[eSIRS]{  \includegraphics[scale=0.23]{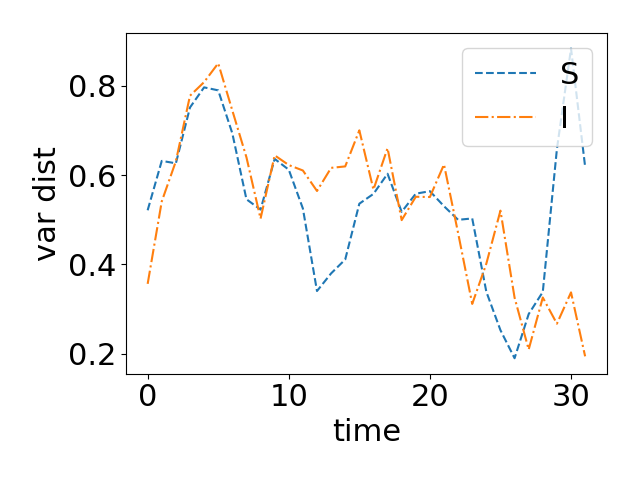}}
    \subfigure[eSIRS-1P]{  
    \includegraphics[scale=0.23]{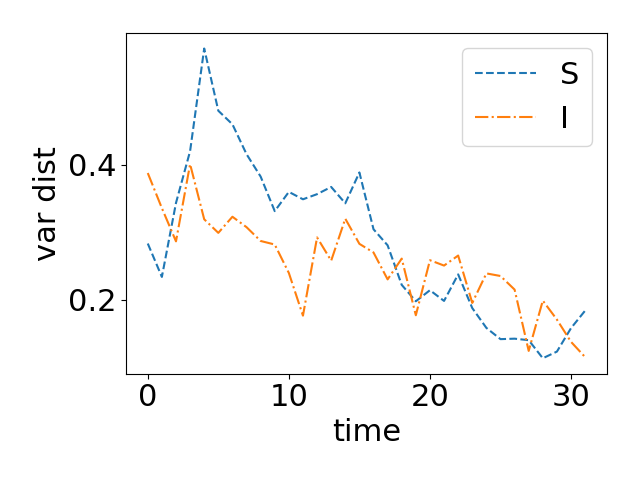}}
    \subfigure[SIR]{  
   \includegraphics[scale=0.23]{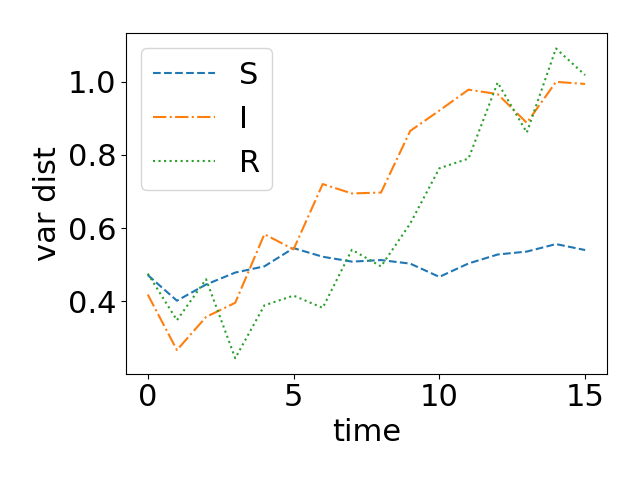}}
    \subfigure[Oscillator]{  
    \includegraphics[scale=0.23]{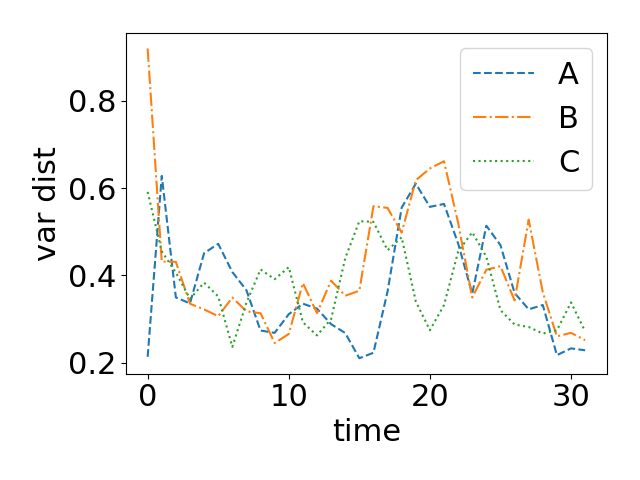}}
    \subfigure[Toggle Switch]{  
    \includegraphics[scale=0.23]{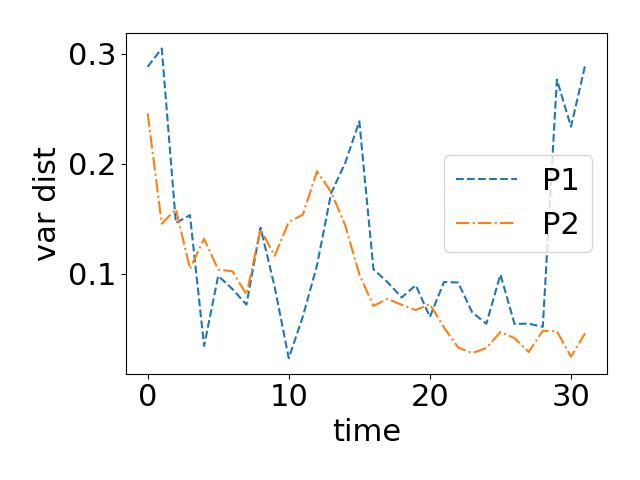}}
     \subfigure[MAPK]{  
    \includegraphics[scale=0.23]{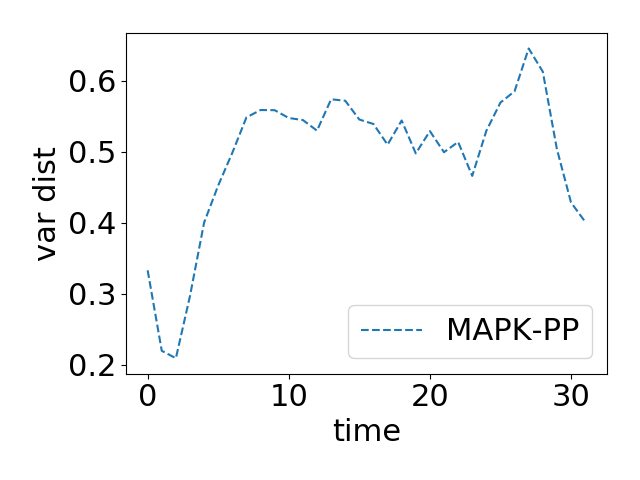}}

    \caption{Plots of the average relative difference in the variances over time for each model and each species. Errors are computed over the entire test set. Generated trajectories have been keep scaled back to $\mathbb{N}$.}\label{fig:avg_vars_rel_errors}
\end{figure}

\begin{figure}[ht]
    \centering
    
    \subfigure[Means abs. err.]{
    \includegraphics[scale=0.12]{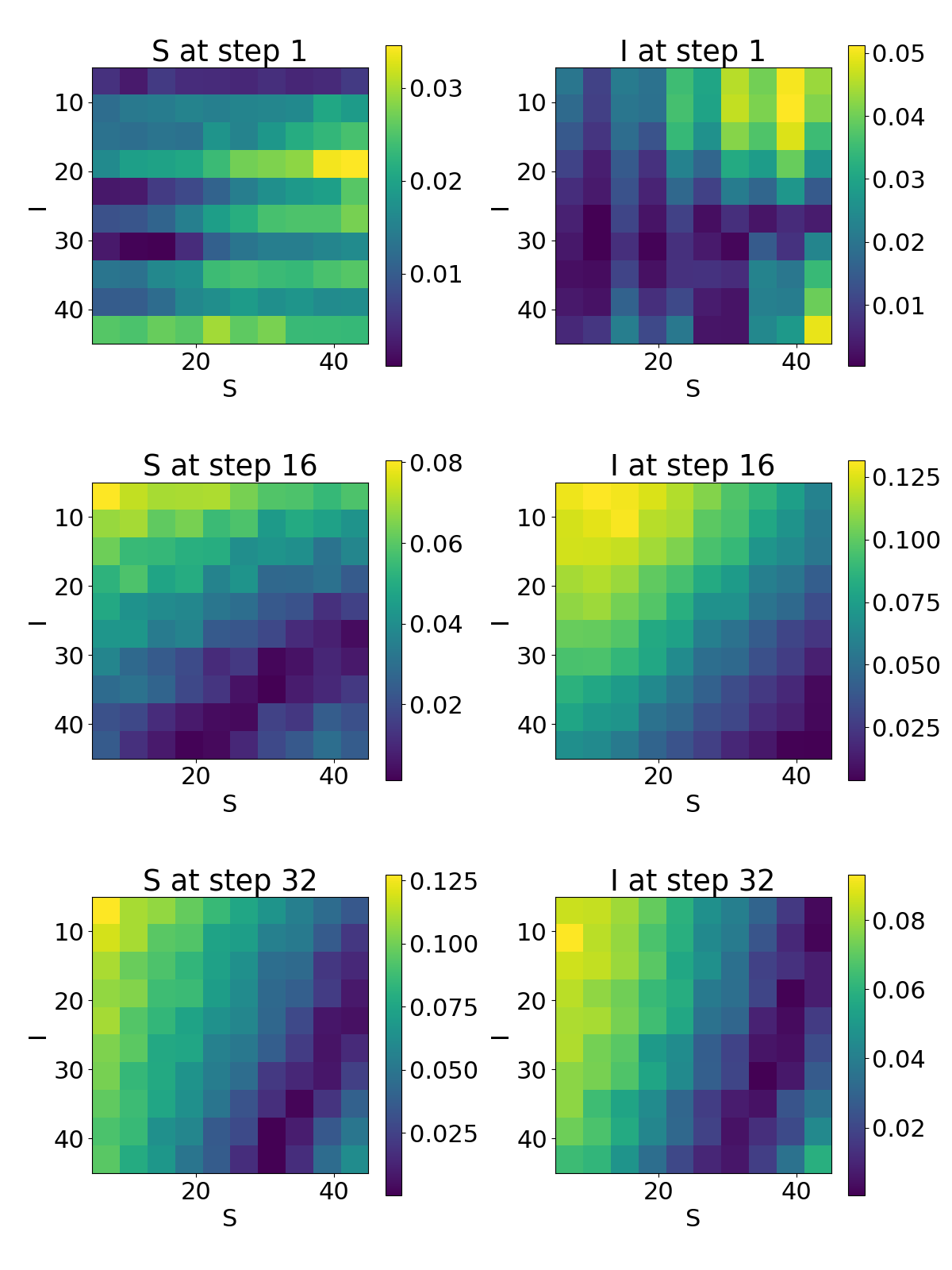}}
        \subfigure[Means rel. err.]{
    \includegraphics[scale=0.12]{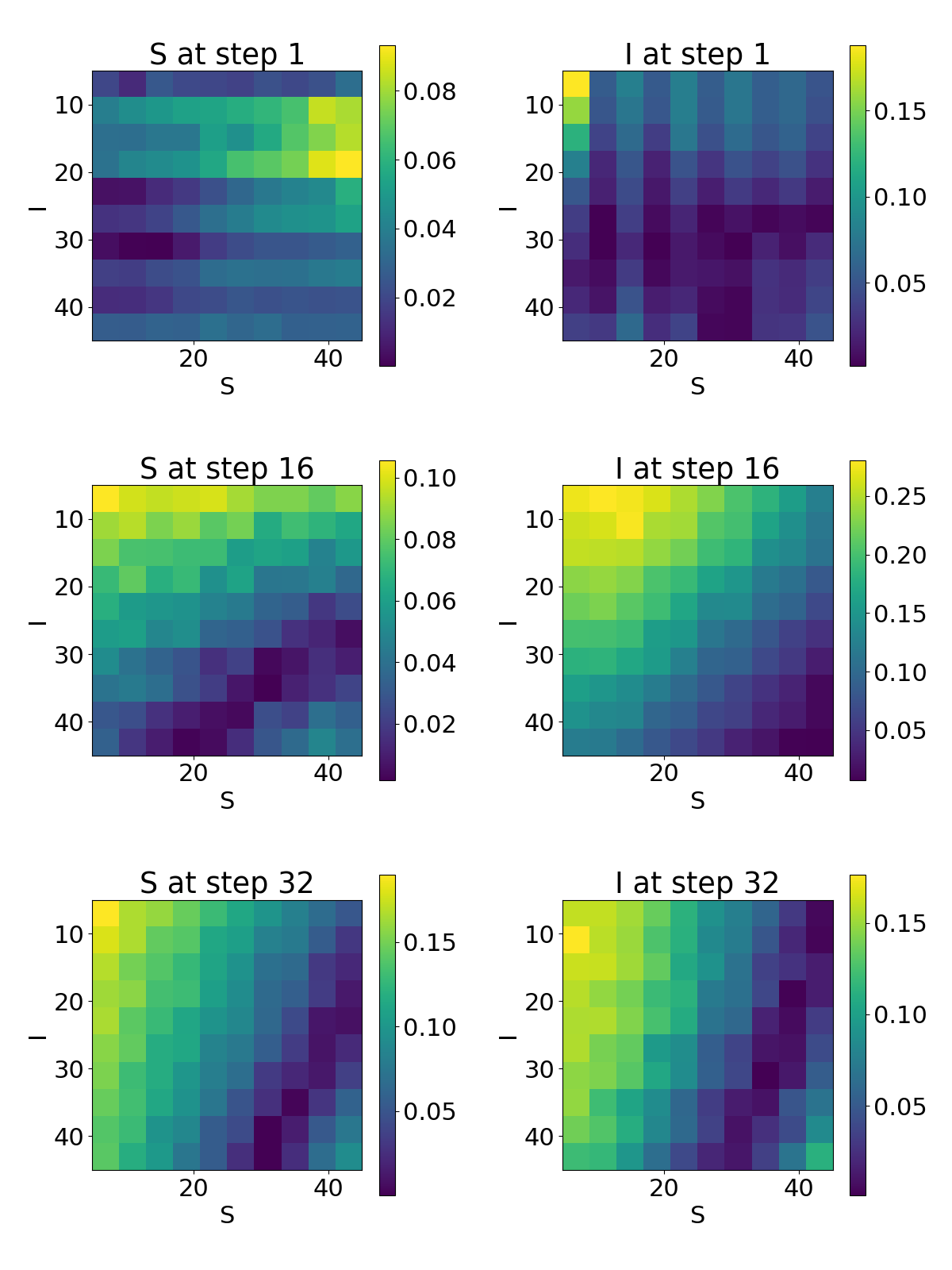}}
    \subfigure[Variances abs. err.]{
    \includegraphics[scale=0.12]{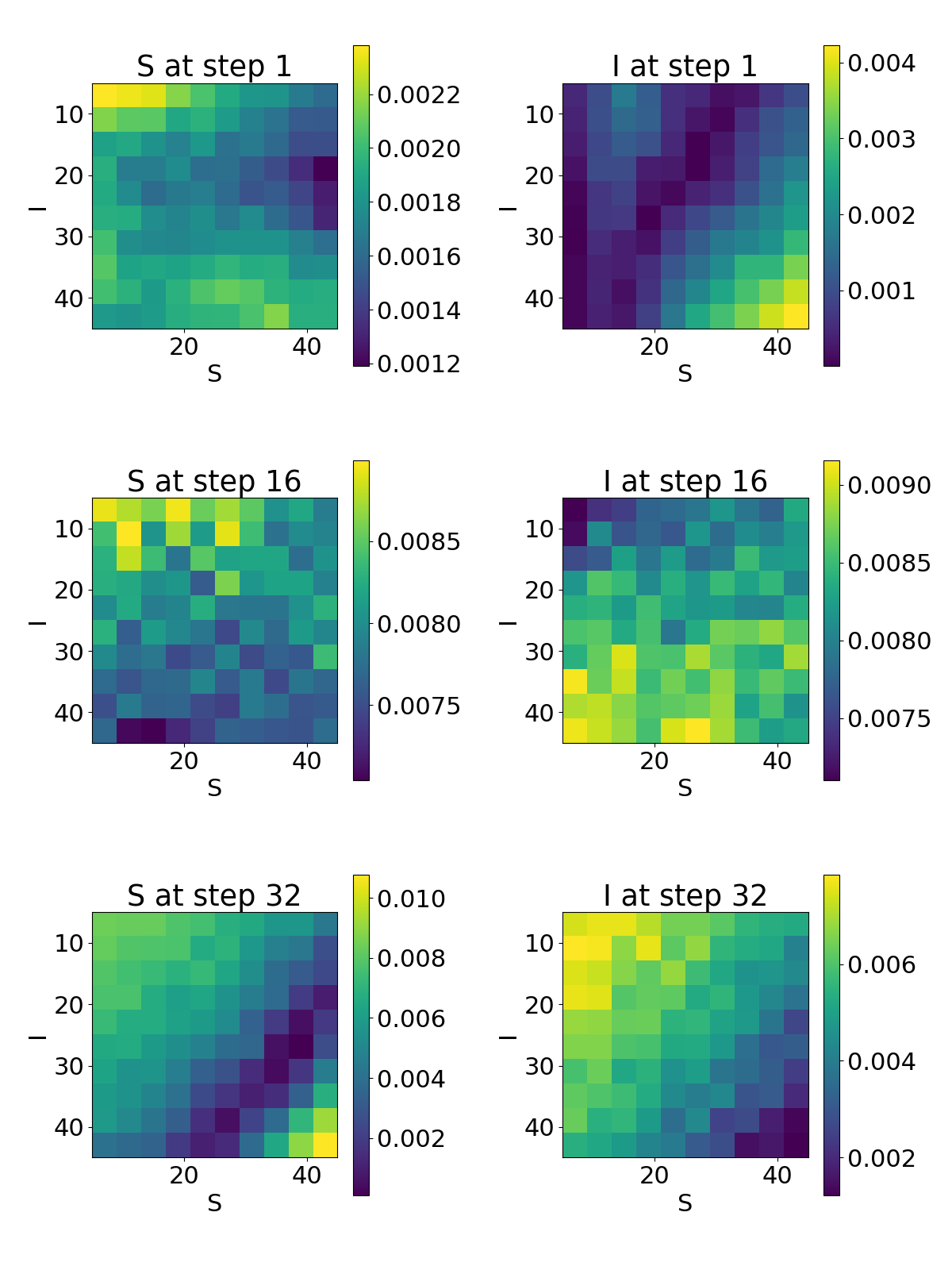}}
    \subfigure[Variances rel. err.]{
    \includegraphics[scale=0.12]{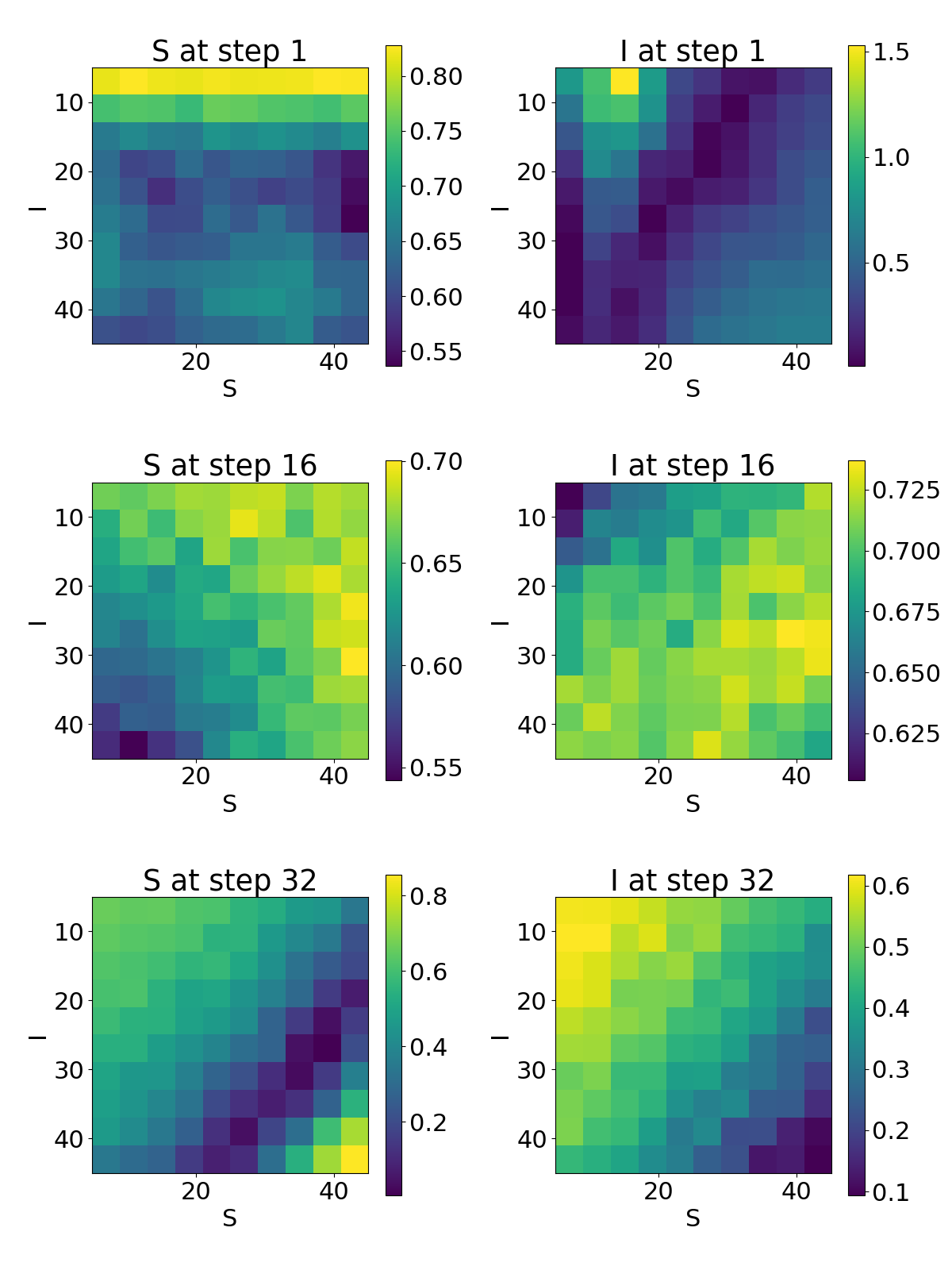}}
    \subfigure[Wass. dist.]{
    \includegraphics[scale=0.12]{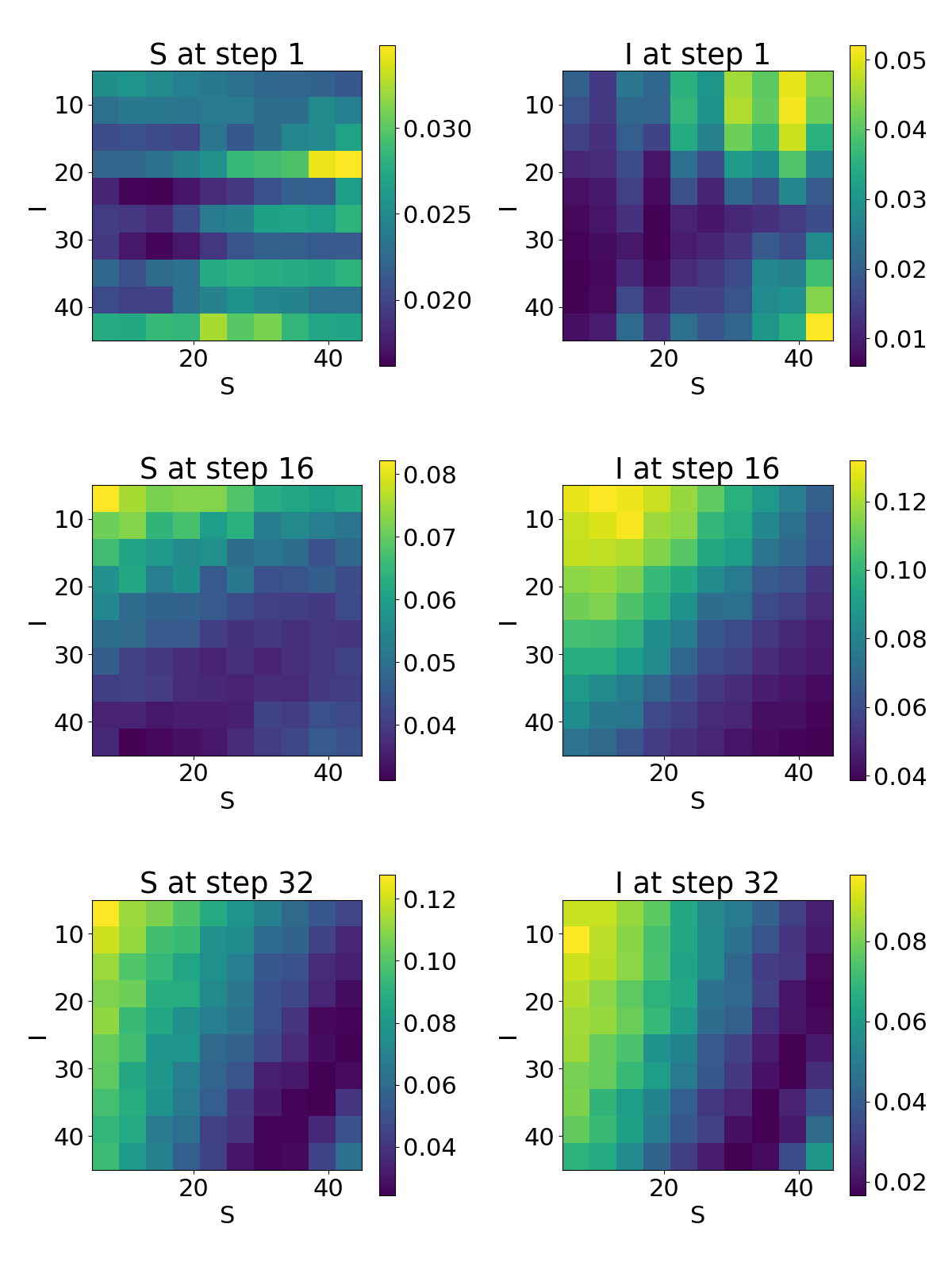}}
    \caption{Histogram distance landscapes for the two-dimensional \textbf{eSIRS} model.}
    \label{fig:eSIRS_distance_landscapes}
\end{figure}

\begin{figure}[ht]
    \centering
    
    \subfigure[Means abs. err.]{
    \includegraphics[scale=0.12]{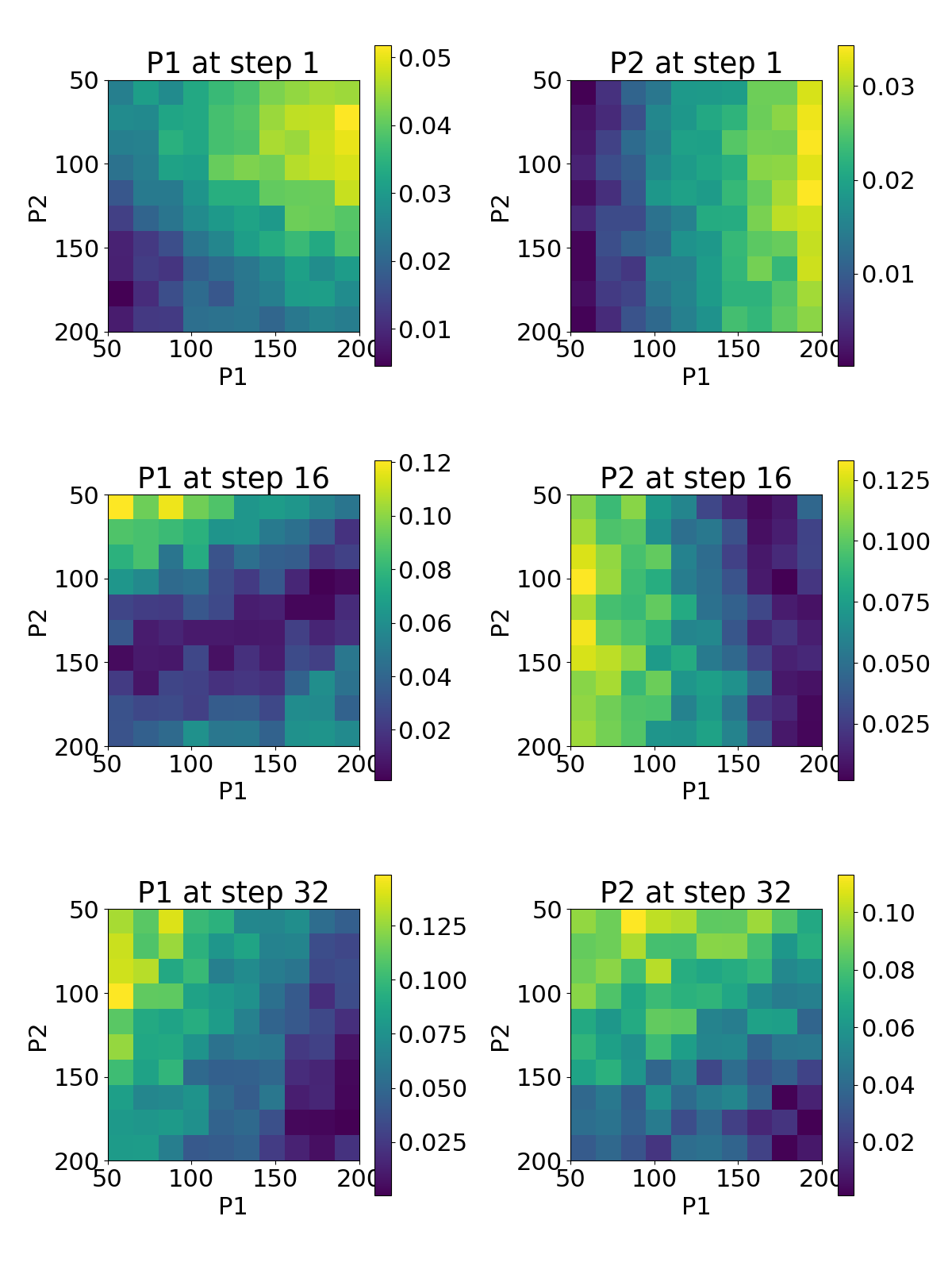}}
            \subfigure[Means rel. err.]{
    \includegraphics[scale=0.12]{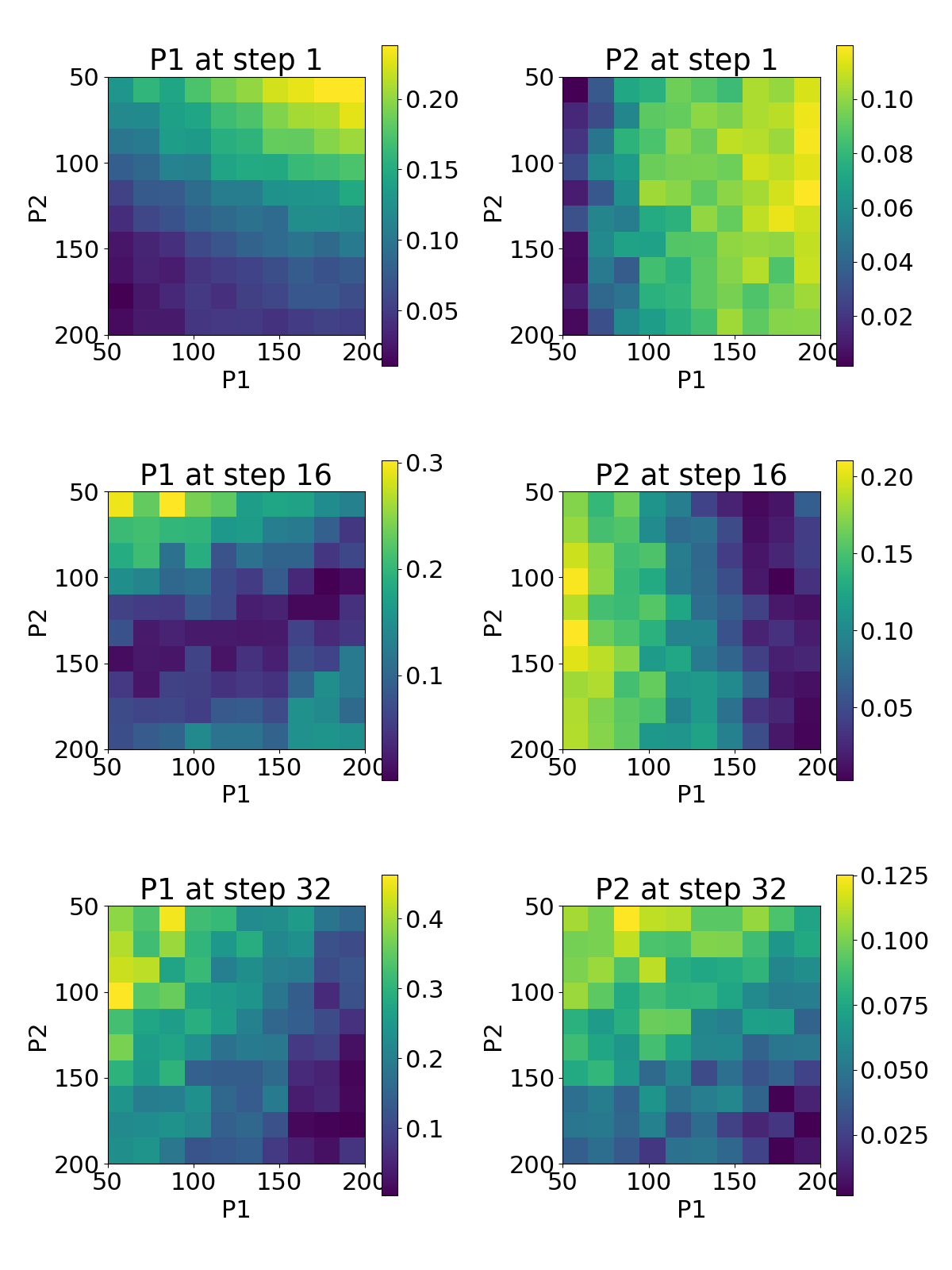}}
    \subfigure[Variances abs. err.]{
    \includegraphics[scale=0.12]{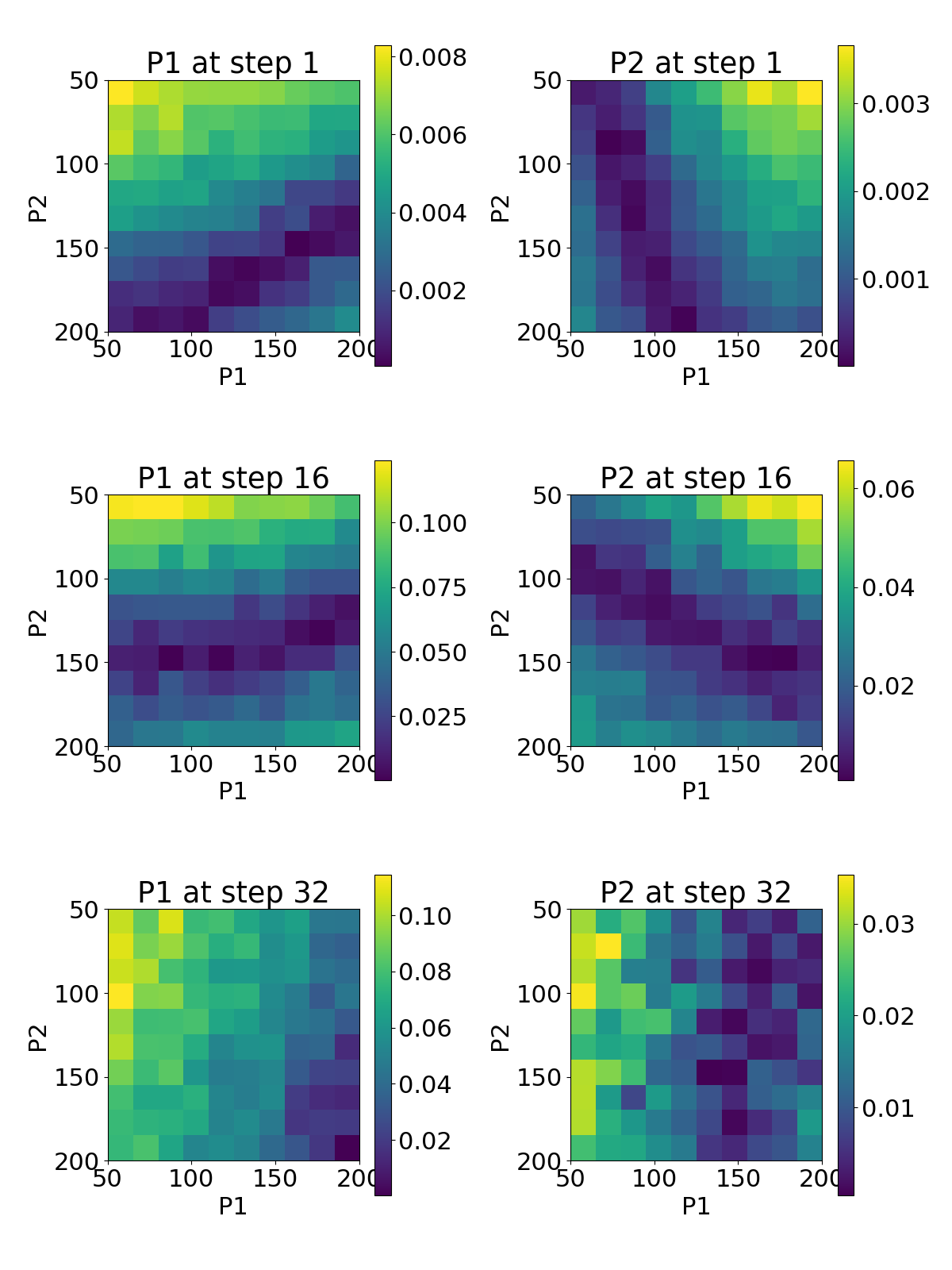}}
    \subfigure[Variances rel. err.]{
    \includegraphics[scale=0.12]{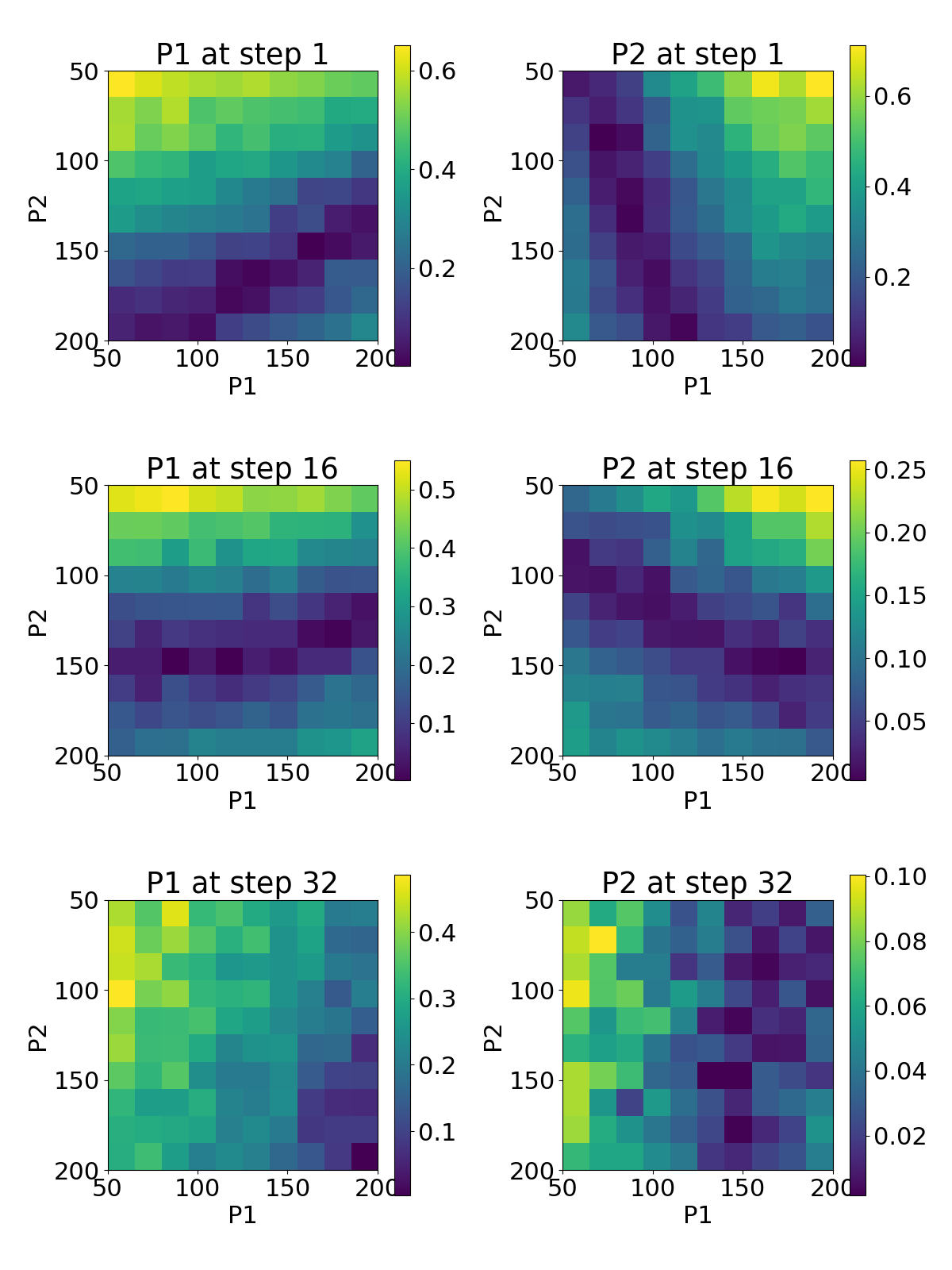}}
    \subfigure[Wass. dist.]{
    \includegraphics[scale=0.12]{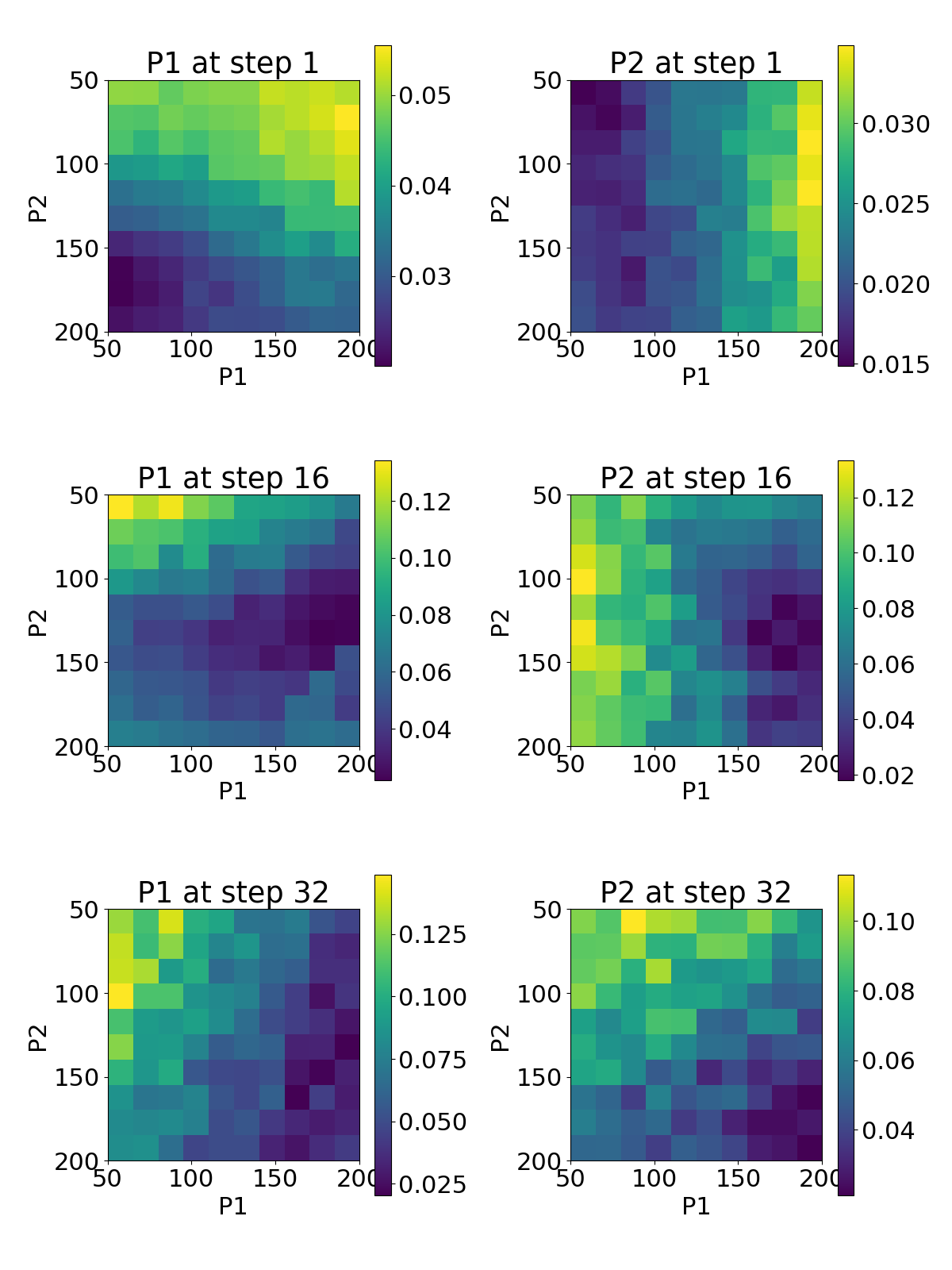}}
    \caption{Histogram distance landscapes for the two-dimensional \textbf{Toggle Switch} model.}
    \label{fig:TS_distance_landscapes}
\end{figure}

\begin{figure}[ht]
    \centering
    \subfigure[Wass. dist.]{
    \includegraphics[scale = 0.24]{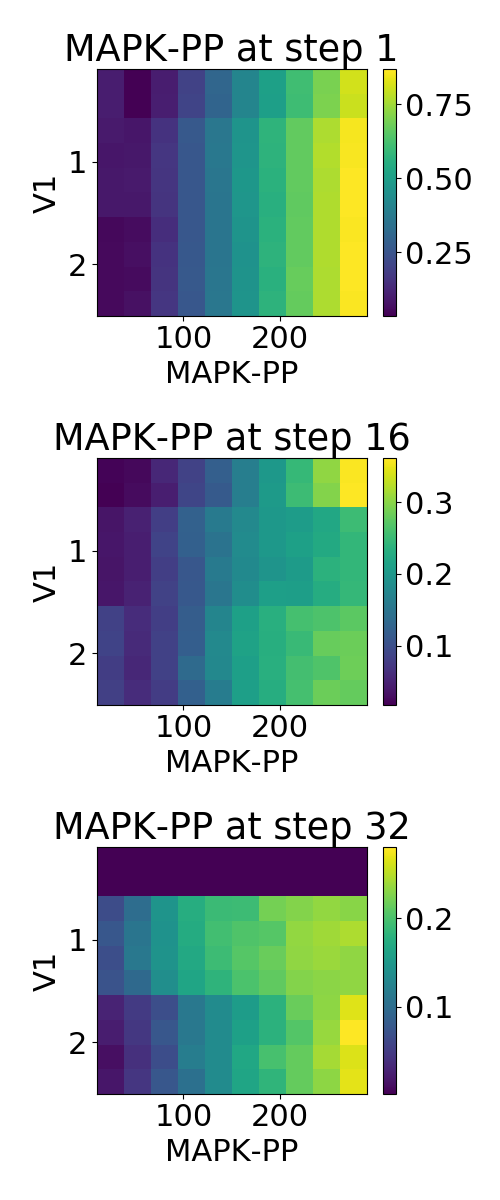}}
    \subfigure[Means abs. err.]{
    \includegraphics[scale = 0.24]{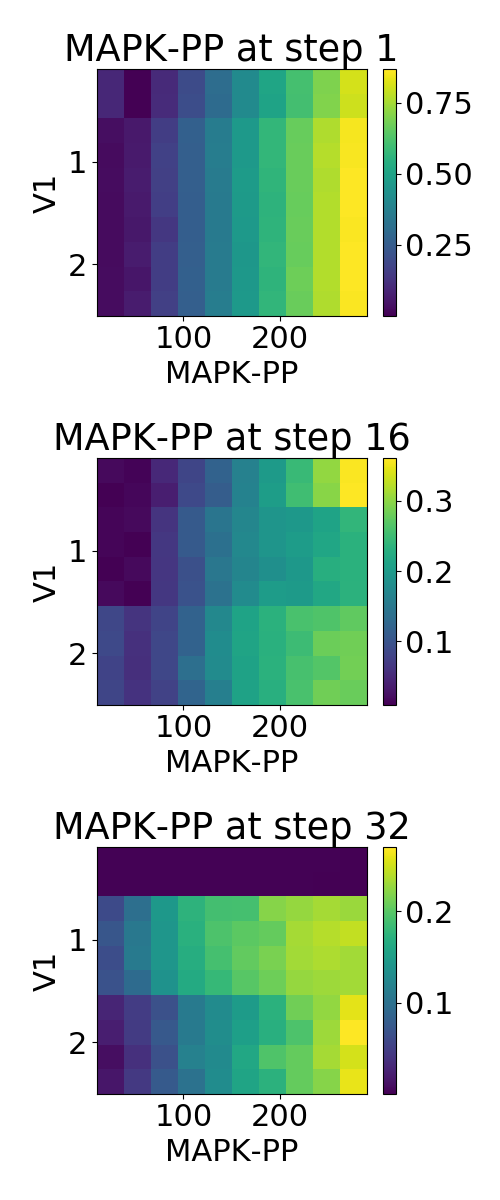}}
    \subfigure[Means rel. err.]{
    \includegraphics[scale = 0.24]{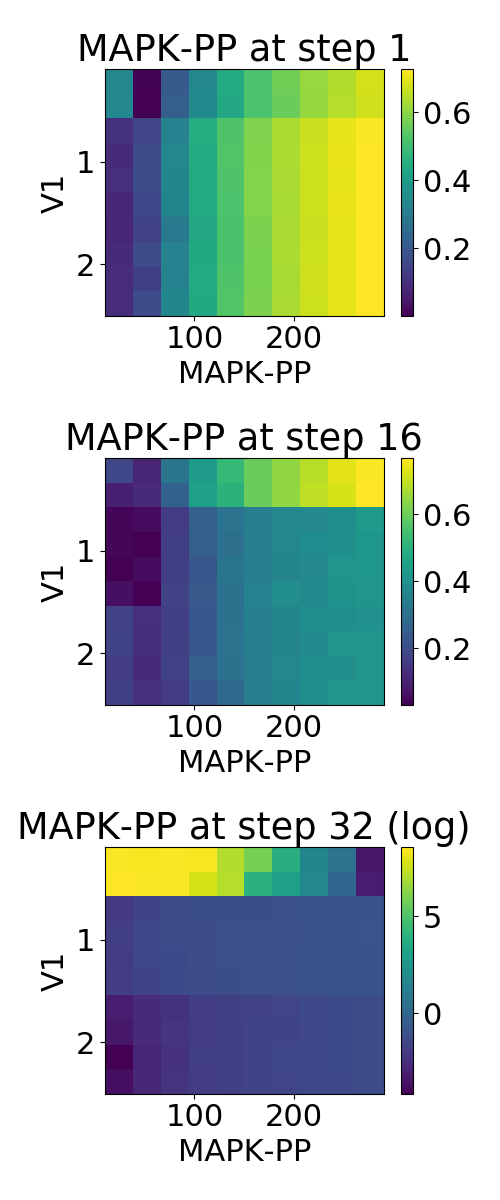}}
    \subfigure[Variances abs. err.]{
    \includegraphics[scale = 0.24]{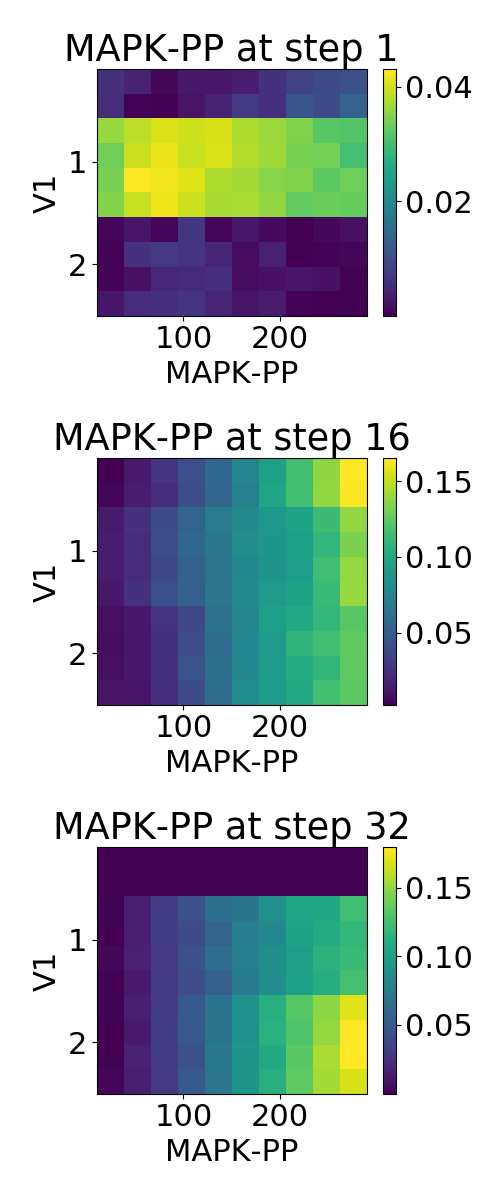}}
    \subfigure[Variances rel. err.]{
    \includegraphics[scale = 0.24]{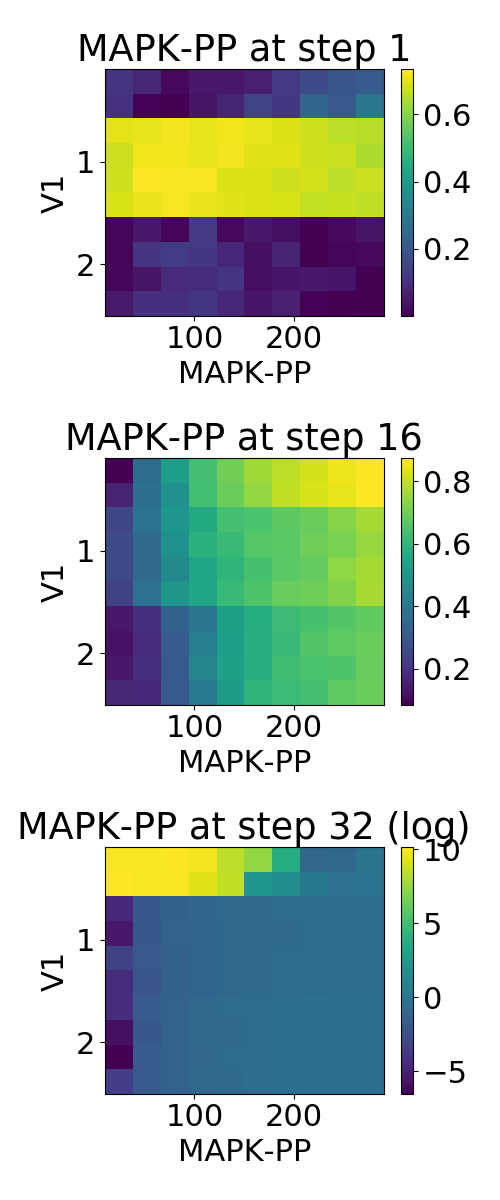}}
    
    \caption{Histogram distance landscapes for the \textbf{MAPK} model.}
    \label{fig:mapk_distance_landscapes}
\end{figure}

\begin{figure}
    \centering
    \includegraphics[scale=0.245]{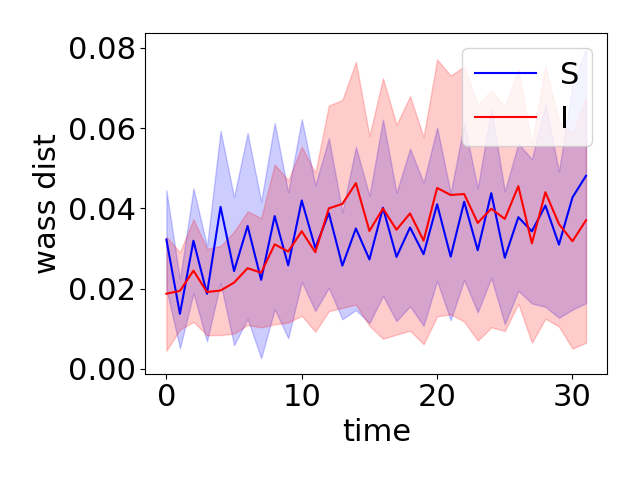}
    \includegraphics[scale=0.245]{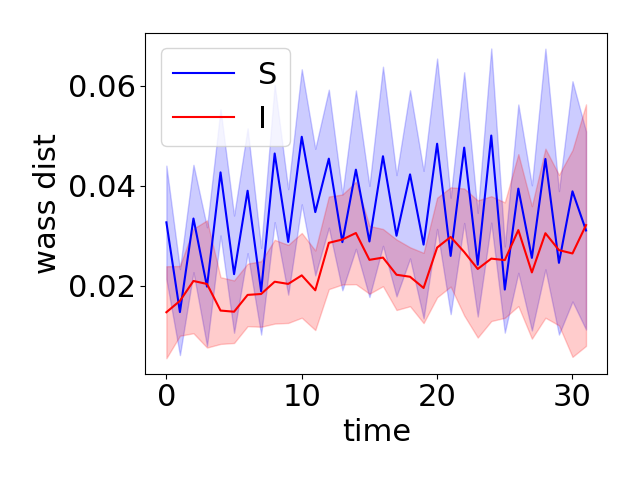}
    \includegraphics[scale=0.245]{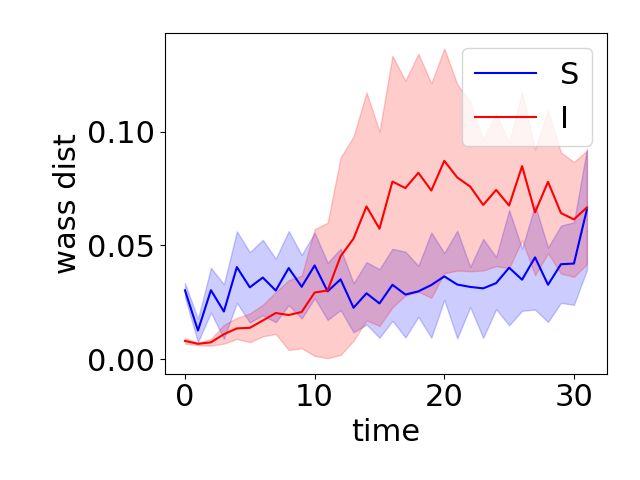}
    \caption{Analysis of the generalization capabilities of the abstract model on various test sets: 100 different pairs $(s_0, \theta)$ \textbf{(left)}, fixed parameter and 100 different initial states \textbf{(middle)} and a fixed initial state with 100 different parameters \textbf{(right)}. For each test set we compute the mean  and the standard deviation of the distribution of Wasserstein distances over such sets.}
    \label{fig:esirs_1p_analysis}
\end{figure}

\section{Satisfaction probability}\label{sec:satisf}
\begin{figure}[ht]
    \centering
    \includegraphics[scale=0.17]{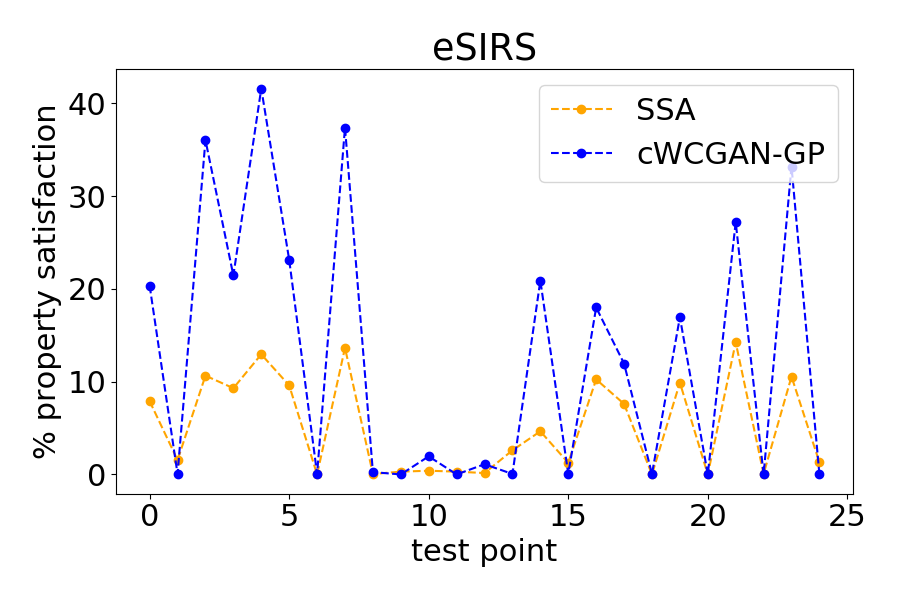}
     \includegraphics[scale=0.17]{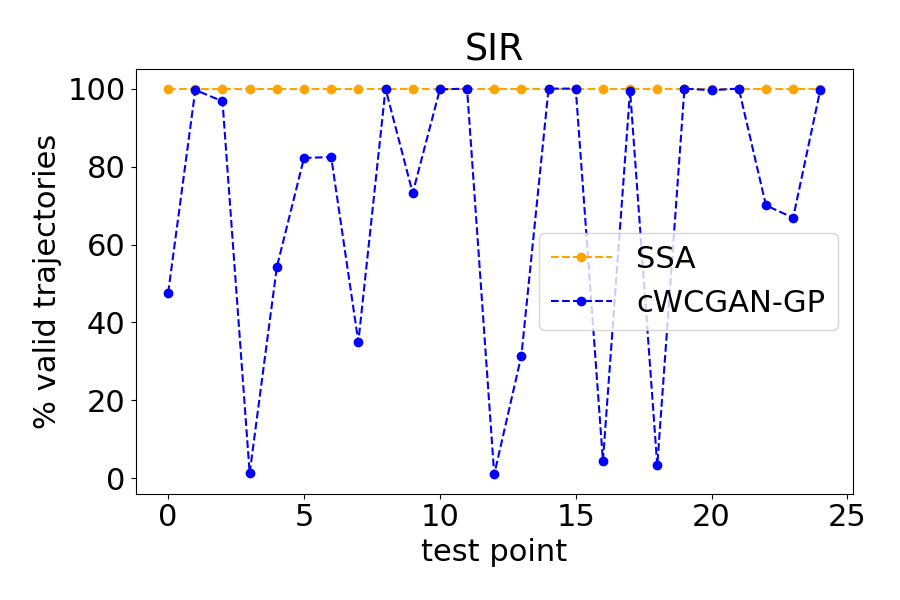}
     \includegraphics[scale=0.17]{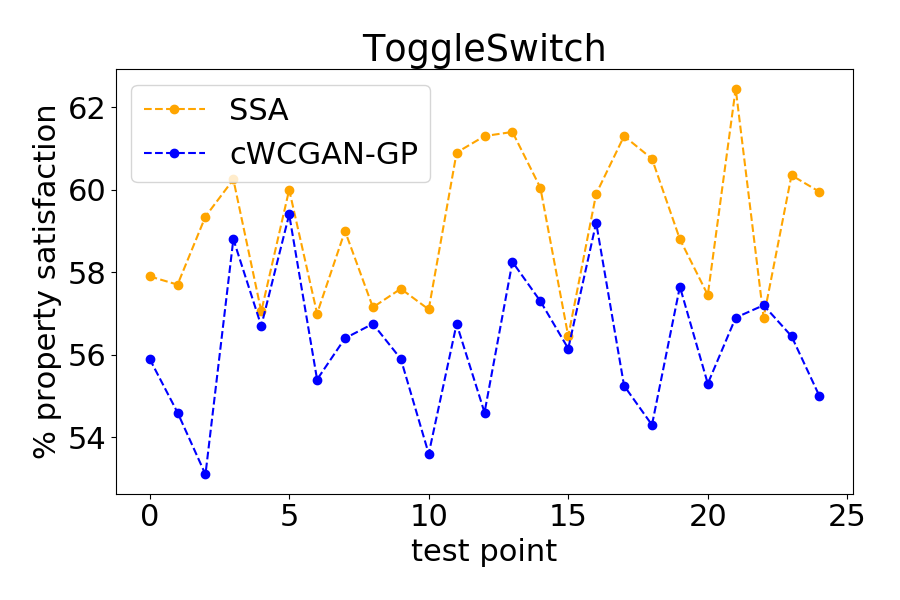}
    \caption{\textbf{(eSIRS)} Given the property ``eventually the number of infected stays below a threshold of 25 individual", we check for each test point (x axis) the percentage of SSA (orange) and abstract (blue) trajectories that satisfy such property.
    \textbf{(SIR)} For abstract trajectories of the SIR model we check, for each test point, the percentage of valid trajectories, i.e. such that the state $I=0$ is absorbing.
    \textbf{(Toggle Switch)} Given the property ``eventually the level of protein P2 stays above a threshold of 50", we check for each test point (x axis) the percentage of SSA (orange) and abstract (blue) trajectories that satisfy such property.
    \vspace{-0.25cm}}
    \label{fig:satisf}
\end{figure}

We seek a formal way to quantify whether the abstract model captures and preserves the emergent macroscopic behaviors of the original system. In order to do so, we can resort to formal languages, such as Signal Temporal Logic (STL) \cite{maler2004monitoring}. The first step is to express formally the property that we would like abstract trajectories to preserve. Then we can measure the satisfaction probability of such property for both real and abstract trajectories and check if it is similar on a large pool of initial settings. Examples are shown in Fig. \ref{fig:satisf}. For the e-SIRS model we consider the property ``eventually the number of infected remains below a threshold of 25 individual". For abstract trajectories of the SIR model we check, for each test point, the percentage of valid trajectories, i.e. such that the state $I=0$ is absorbing. Finally,  for the Toggle Switch model we check the property ``eventually the level of protein $P_2$ stays above a threshold of $50$", meaning we check for each test point the percentage of SSA and abstract trajectories that satisfy such property. It can be written as $\Diamond_{[0,H]}\square (P_2 > 50)$. These comparisons produce a measurable qualitative estimate of how good the reconstruction is. As future work, we intend to use such qualitative measure as a query strategy for an active learning approach, so that the obtained abstract model is driven in the desired direction.

\section{Statistical tests}\label{sec:stat_test}
\begin{figure}[ht]
    \centering
    \includegraphics[scale=0.245]{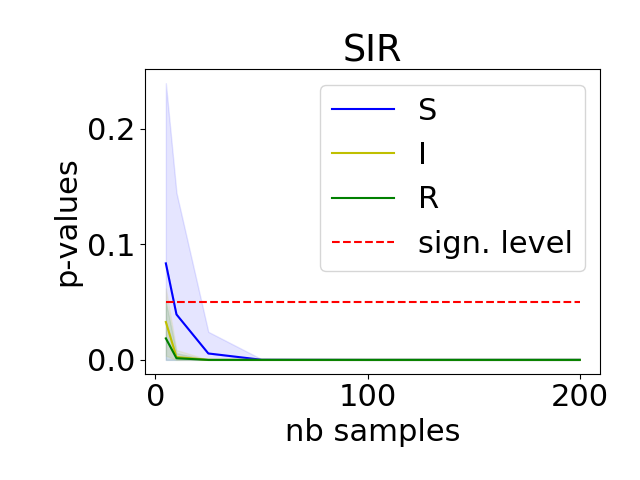}
    \includegraphics[scale=0.245]{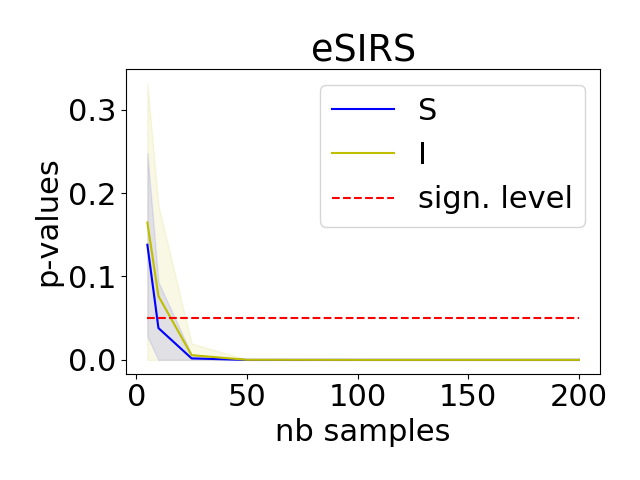}
    \includegraphics[scale=0.245]{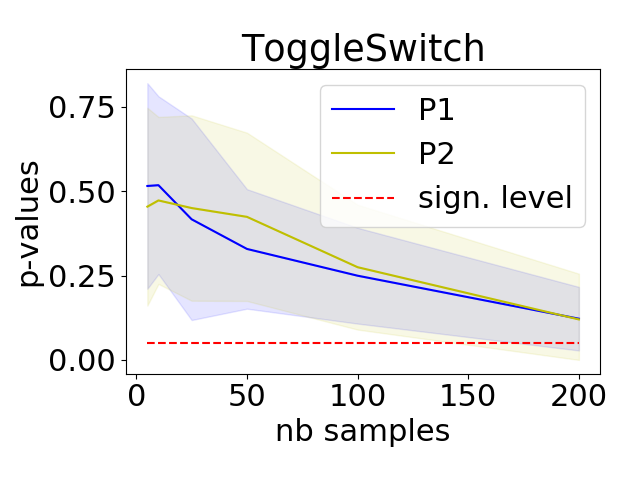}
    \includegraphics[scale=0.245]{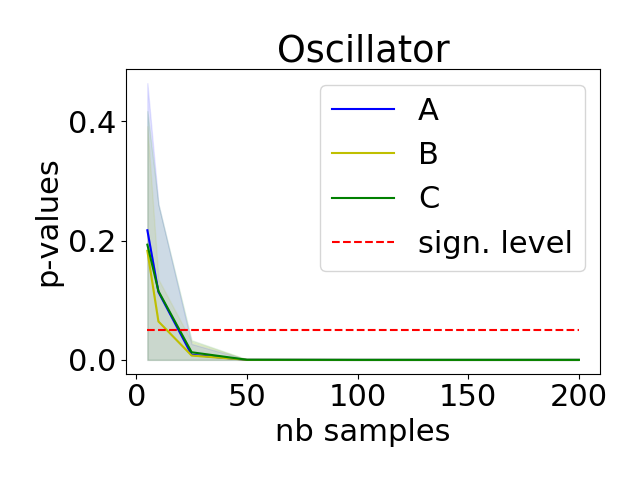}
    \includegraphics[scale=0.245]{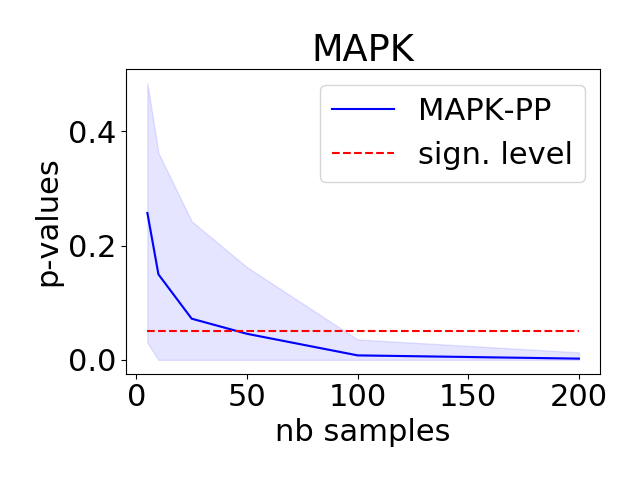}
    \caption{Average over the initial setting of the p-values (with confidence interval) for each species computing by the two-sample statistical test w.r.t. the number of samples present in the empirical distributions.}
    \label{fig:stat_test}
\end{figure}
In this section we show the results of a two-sample statistical test over all the cases studies. In particular, we use a statistical test based on the energy distance among distributions \cite{szekely2013energy}. For each initial setting and for each species we compute the distance statistics and the p-value among the empirical approximation of the SSA distribution and the empirical approximation of the abstract distribution over the trajectory space, i.e. a $H$-dimensional space. In Fig. \ref{fig:stat_test} we report the mean and the standard deviation of p-values over the initial settings present in the test set. Clearly the p-value decreases as the number of samples used to approximate the distributions increases. Fig. \ref{fig:stat_test} shows how the p-values for each species varies according to the number of samples used.
These results come with no surprise as the abstract model was trained having only $10$ observations for each initial setting. It is interesting to observe how the Energy test is passed by a large percentage of points when the number of samples is around $10$. In order to enhance the resilience of the abstract model to such statistical tests we should increase the number of samples per point in the training set. This comes at the cost of reducing the number of initial setting, so that the resulting training set is not too large. In this regard, the active learning technique proposed in Section \ref{sec:satisf} can be extremely beneficial.

\section{cWCGAN-GP Algorithm}\label{sec:algorithm}

\begin{algorithm}[ht]

 \KwData{The gradient penalty coefficient $\lambda$, the number of epochs $n_{epochs}$, the number of critic iterations per generator iteration $n_{critic}$, the batch size $m$, Adam hyper-parameters $\alpha, \beta_1, \beta_2$.}
 \For{$e=1\dots, n_{epochs}$}{
  \For{$t=1\dots, n_{critic}$}{
  \For{$i=1\dots, m$}{
        Sample real data $(y, x)\sim P_r$, latent variable $z \sim p(z)$, a random number $\epsilon\sim U[0, 1]$\;
        $\tilde{x}\leftarrow G_{w_g}(z, y)$\;
        
        $\hat{x}\leftarrow \epsilon x +(1-\epsilon)\tilde{x}$\;
        $L^{(i)}\leftarrow D_{w_c}(\tilde{x}, y)-D_{w_c}(x, y)+\lambda(\parallel \nabla_{\hat{x}} D_{w_c}(\hat{x},y)\parallel_2-1)^2$\; 
        }
        {
        $w_c\leftarrow$ Adam $\left(\nabla_{w_c} \tfrac{1}{m}\sum_{i=1}^m L^{(i)}, w_c, \alpha, \beta_1, \beta_2\right)$\;
        }
   }
   {
   Sample a batch of latent variables $\{z^{(i)}\}_{i=1}^m\sim p(z)$\ and a batch of random conditions $\{y^{(i)}\}_{i=1}^m\sim p(y)$\;
   $\theta\leftarrow $ Adam $\left(\nabla_{w_g} \tfrac{1}{m}\sum_{i=1}^m -D_{w_c}(G_{w_g}(z^{(i)}, y^{(i)}), y^{(i)}), w_g, \alpha, \beta_1, \beta_2 \right)$
   }
   
   {}
 }
 \caption{Conditional WGAN with gradient penalty. Default values used for hyper-parameters: $\lambda = 10$, $n_{critic} = 5$, $\alpha = 0.0001$, $\beta_1 = 0.5$, $\beta_2 = 0.9$. Variable $x$ denotes the trajectories of length $H$ ($\eta_{[1,H]}$), whereas variable $y$ denotes the condition, i.e., the initial setting of the system (pairs $(s_0, \theta)$. }
\end{algorithm}

\end{document}